\documentclass[a4paper,fleqn]{cas-sc}

\usepackage[authoryear]{natbib}
\usepackage{amsmath,amssymb,amsthm,bm}
\usepackage{booktabs,multirow,tabularx,array,makecell}
\usepackage{graphicx}
\graphicspath{{./}{figures/}}
\usepackage{placeins}
\usepackage{microtype}
\usepackage{needspace}
\usepackage{xurl}
\usepackage{float}
\usepackage{tikz}
\usepackage{arydshln}

\usetikzlibrary{arrows.meta,positioning}
\usepackage{algorithm}
\usepackage{algpseudocode}
\newtheorem{lemma}{Lemma}
\newtheorem{theorem}{Theorem}
\newif\ifanonymous

\anonymousfalse

\RenewDocumentCommand\printorcid{}{}
\begin{document}

\shorttitle{Compositional Conformal Risk Control for Multi-Stage LLM Cascades}
\shortauthors{Sun et al.}

\title[mode=title]{Compositional Conformal Risk Control for Multi-Stage LLM Cascades}

\ifanonymous
\shortauthors{Anonymous Author(s)}
\author{Anonymous Author(s)}
\else
\author[]{Yuxia Sun}
\ead{tyxsun@email.jnu.edu.cn}

\author[]{Jiahuang Wang}
\ead{wjhrobbia@stu2025.jnu.edu.cn}

\author[]{Haoting Liu}
\ead{lhting@stu2026.jnu.edu.cn}

\author[]{Zhiwei Huang}
\ead{hzw66@stu2025.jnu.edu.cn}

\author[]{Tingrui Pei}
\ead{tingruipei@jnu.edu.cn}

\author[]{Saiqin Long}
\cormark[1]
\ead{saiqinlong@jnu.edu.cn}

\affiliation[]{
    organization={College of Information Science and Technology, Jinan University},
    city={Guangzhou},
    postcode={510632},
    state={Guangdong},
    country={China}
}

\cortext[1]{}
\fi

\begin{abstract}
Multi-stage large language model (LLM) cascades balance accuracy and inference cost, but reliable deployment also requires controlling error risk and identifying its sources across stages. We propose Compositional Conformal Risk Control (CompCRC), a risk-control framework for pre-specified multi-stage LLM cascades. We decompose the risk of accepting incorrect answers into additive stage-wise contributions over the full query population, linking local risk budgets to a system-level guarantee. CompCRC constructs and freezes candidate budget allocations and acceptance thresholds, selects one candidate on independent data, and certifies its stage-wise risk contributions on separate held-out data. Under conditional exchangeability, we establish a finite-sample guarantee: the marginal expected accepted-error risk of the complete deployment procedure does not exceed the global risk budget. At the reported comparison budget of $\alpha=0.45$, all three stages of CompCRC pass their local checks on MMLU,
GSM8K, and HellaSwag. Compared with Global-CRC, which passes system-level risk certification at the same budget, CompCRC reduces observed accepted-error risk by 1.33--2.00 percentage points, improves final accuracy by 0.67--2.00 percentage points, and reduces weighted inference cost by 30.37\% on HellaSwag. The selected cost-based budget allocation also improves accuracy per unit cost by 10.6\%--46.7\% over uniform allocation. Further experiments show that stage-wise statistics identify stages that fail local certification checks, providing diagnostic information alongside system-level risk control.
\end{abstract}



\maketitle

\section{Introduction}
\label{sec:introduction}

Practical deployment of large language models (LLMs) requires balancing
answer accuracy and inference cost. Model cascading and routing address
this trade-off by selectively invoking models with different costs and
capabilities. A multi-stage cascade typically starts with a lower-cost
model and either returns its answer if an acceptance criterion is met
or forwards the query to a subsequent model.
Existing studies show that selectively invoking models based on query
characteristics or estimated answer quality can improve the
accuracy--cost trade-off
\citep{Chen2020FrugalML,Chen2024FrugalGPT,Ong2025RouteLLM}.
However, when deployment requires a predefined upper bound on error
risk, strong test-set performance alone does not provide the
corresponding statistical guarantee. Establishing such a guarantee from
finite data while balancing accuracy and inference cost is therefore
an important challenge for reliable cascade inference.

Conformal risk control (CRC) provides a theoretical foundation for
addressing this challenge by giving finite-sample expected-risk
guarantees for bounded monotone losses under exchangeability
\citep{Angelopoulos2024CRC}. Recent studies apply conformal calibration
to routing thresholds, model selection, and prediction-set construction
in model cascades, supporting statistically reliable adaptive inference
\citep{Dou2026ConformalCascade,Guo2026RouteNLP,Xue2025ConformalRouting}.
However, a system-level risk guarantee alone does not quantify
each stage's contribution to the overall error risk. Furthermore, supporting
stage-wise risk control and diagnosis requires linking the
system-level risk requirement to individual stages, quantifying their
risk contributions, and identifying stages that do not meet local
risk requirements.

Our central idea is to decompose accepted-error risk---the probability
of accepting an incorrect answer for a randomly drawn query---across
cascade stages. A stage's risk contribution depends on both the error
rate among its accepted answers and the fraction of all queries it
accepts. Because each query can be accepted by at most one stage, these
contributions sum exactly to the system-level accepted-error risk.
This relationship allows a global risk budget to be allocated across
stages, connecting local risk control to a system-level risk guarantee,
while supporting stage-wise risk analysis and diagnosis.

Building on this idea, we propose Compositional Conformal Risk Control
(CompCRC) for a pre-specified multi-stage LLM cascade.
CompCRC first constructs and freezes candidate allocations of the global
risk budget and their corresponding acceptance thresholds.
It then uses independent data to select a candidate that balances
accuracy and cost.
Each stage's risk contribution is subsequently certified on a dedicated
held-out set disjoint from the data used for candidate construction
and selection.
The selected cascade is enabled only if all stages pass certification;
otherwise, a pre-specified certification-failure fallback policy is used.
Under conditional exchangeability, we establish a finite-sample guarantee:
the marginal expected accepted-error risk of this complete deployment
procedure does not exceed the global risk budget.
By comparing each stage's certification statistic with its local risk
budget, CompCRC also identifies stages that fail certification, providing
diagnostic information alongside system-level risk control.

We evaluate CompCRC on MMLU, GSM8K, and HellaSwag using a three-model
cascade.
At the reported comparison budget of $\alpha=0.45$,
CompCRC passes local certification at every stage on all
three benchmarks.
Compared with Global-CRC, which passes system-level risk certification
at the same budget, CompCRC reduces observed accepted-error risk by
1.33, 1.67, and 2.00 percentage points and improves final accuracy by
1.33, 0.67, and 2.00 percentage points, respectively.
On HellaSwag, CompCRC also reduces weighted inference cost by 30.37\%.
Compared with uniform budget allocation, the selected cost-based budget
allocation improves accuracy per unit cost by approximately
16.8\%, 10.6\%, and 46.7\% on the three benchmarks, respectively.
Budget-sensitivity experiments further show that stage-wise certification
identifies stages that fail local risk checks.

Our main contributions are as follows:
\begin{itemize}
    \item \textbf{CompCRC risk-control framework.}
    We propose a framework that combines system-level risk control
    with diagnosis of local certification failures in multi-stage
    LLM cascades, and uses budget allocation to balance accuracy
    and cost.

    \item \textbf{Risk decomposition and statistical guarantee.}
    We establish an additive decomposition of system-level
    accepted-error risk into stage-wise contributions.
    Under conditional exchangeability, we provide a finite-sample
    guarantee that the marginal expected accepted-error risk of
    the complete CompCRC deployment procedure does not exceed
    the global risk budget.

   \item \textbf{Empirical evaluation of performance and risk diagnosis.} Experiments show that, under the same global risk budget, CompCRC achieves lower observed accepted-error risk and higher final accuracy than Global-CRC; the selected budget allocation also improves accuracy per unit cost over uniform allocation. Further experiments demonstrate CompCRC's stage-wise risk diagnosis capability.
\end{itemize}

\section{Related Work}
\label{sec:related_work}

\subsection{LLM Routing and Multi-Stage Cascades}
\label{subsec:routing_cascades}

Model routing and cascading reduce inference cost by adapting model use to query difficulty. Routing typically selects a model for a query, whereas cascading can invoke additional models after an initial response. In a cascade, the cost of answering a query includes every invoked model, making the decision to continue consequential for both answer quality and total computation. FrugalML and early NLP cascades combine models with different costs to improve prediction efficiency \citep{Chen2020FrugalML,Varshney2022ModelCascading}, while FrugalGPT extends this approach to LLM services \citep{Chen2024FrugalGPT}. RouteLLM learns routing decisions from preference data, HybridLLM studies quality-aware routing between models, and TensorOpera Router supports multi-model inference \citep{Ong2025RouteLLM,Ding2024HybridLLM,Stripelis2024TORouter}.

Complementary studies examine the signals needed for cascade deferral: \citet{Gupta2024LanguageCascades} analyze token-level uncertainty, and \citet{Jitkrittum2023ConfidenceCascade} characterize when confidence-based deferral suffices. These analyses connect deferral decisions to the current model's confidence and the potential benefits of downstream inference. This distinction matters when downstream models have specialized strengths and do not improve equally on every input. More recent frameworks combine routing with conformal calibration or calibrated error probabilities. RouteNLP integrates difficulty-aware routing, conformal cascading, and distillation, using conformal prediction to initialize cascade thresholds. Its distillation component uses escalation failures to improve cheaper models, followed by router retraining to reflect the updated model capabilities \citep{Guo2026RouteNLP}. UCCI calibrates uncertainty into per-query error probabilities and uses these estimates for cost-optimal cascade decisions. Under stated assumptions, its analysis establishes the cost optimality of threshold policies based on the calibrated score \citep{Kotte2026UCCI}. 

These studies primarily develop routing and deferral policies. However, routing and deferral policies alone do not provide statistical checks of each stage's accepted-error risk against its allocated budget. They also do not identify which stages fail these checks.

\subsection{Conformal Prediction and Risk Control}
\label{subsec:cp_crc}

Conformal prediction provides finite-sample coverage guarantees for prediction sets under exchangeability \citep{Vovk2005Algorithmic,Shafer2008Tutorial, Angelopoulos2023GentleIntroduction}, with applications throughout natural language processing \citep{Campos2024ConformalNLP}. In split conformal prediction, held-out examples calibrate a score threshold that determines which outputs are included in a prediction set. Risk-controlling prediction sets extend calibration to general risk targets beyond miscoverage \citep{Bates2021RiskControllingSets}, and CRC controls the expected value of bounded losses that are monotone in a calibration parameter \citep{Angelopoulos2024CRC}. These extensions allow calibration objectives to reflect task-specific losses, broadening the range of risks that can be controlled. For language generation, conformal methods construct calibrated sets of candidate outputs, with guarantees on the inclusion of an acceptable response. This can involve calibrating both a sampling stopping rule and a candidate rejection rule to obtain compact sets of responses \citep{Quach2024ConformalLanguageModeling}. Confidence estimation and semantic entropy provide complementary uncertainty signals for assessing the reliability of individual outputs and the variability of possible responses \citep{Lin2024GeneratingConfidence,Farquhar2024SemanticEntropy}.

Two related lines of work apply these ideas directly to adaptive LLM inference. Conformal Cascade constructs stage-wise prediction sets and uses their size to decide whether to accept or defer, accepting only when the calibrated set contains a single answer \citep{Dou2026ConformalCascade}. Its guarantees connect prediction-set calibration to the cascade's acceptance rule, linking further model invocation to uncertainty expressed through the prediction set. Conformal Risk-Controlled Routing calibrates a system-level routing loss involving a small model and a candidate set of larger models \citep{Xue2025ConformalRouting}. 

However, these guarantees alone do not quantify each stage's accepted-error risk contribution over the full query population. Such a contribution accounts jointly for whether a query is accepted at a stage and whether the accepted response is incorrect. These guarantees also leave a question open: once a policy is selected, how can independent stage-wise risk checks ensure risk control for the complete deployment procedure?

\section{Problem Formulation}
\label{sec:problem_formulation}

We study risk-controlled deployment of a pre-specified multi-stage
LLM cascade. Given the cascade and finite labeled data, the objective
is to determine an acceptance policy and a deployment procedure that
meet a prescribed accepted-error risk requirement while balancing
final accuracy and inference cost.

\vspace{0.5\baselineskip}
\noindent\textbf{Cascade setting and acceptance policy.}
Consider a fixed, ordered cascade of $K$ large language
models, $\{f_k\}_{k=1}^{K}$.Let $(X,Y)$ denote a random query and its reference answer,
and let $(x,y)$ denote a particular sample.
For an input $x$ with reference answer $y$, stage
$k\in\{1,\ldots,K\}$ produces an answer $a_k(x)$ and a
confidence score $c_k(x)\in[0,1]$. An \emph{acceptance policy}
assigns a threshold $t_k$ to each stage. 

Execution starts with the
first model. When a query reaches stage $k$, the stage accepts its
answer if $c_k(x)\geq t_k$, returns $a_k(x)$, and terminates the
cascade. Queries that are not accepted proceed to subsequent stages.
If all stages reject a query, a pre-specified output rule selects
the final answer from the available model outputs. Thus, each query
is accepted by at most one stage. Let $\mathrm{reach}_k(x)$ denote
the event that query $x$ reaches stage $k$, which occurs when all
preceding stages have rejected the query. For a fixed acceptance
policy and output rule, let $a_{\mathrm{out}}(x)$ denote the final
answer. With stochastic generation, the probabilities and
expectations below also account for model-output randomness under
fixed generation settings.

\vspace{0.5\baselineskip}
\noindent\textbf{System-level accepted-error risk.}
The \emph{system-level accepted-error risk} is the probability that
a random query receives an incorrect answer accepted by a cascade
stage. This risk is defined over the full query population.

Let $\mathbf{1}\{\cdot\}$ denote an indicator, equal to one when
its condition holds and zero otherwise. Define the \emph{error loss}
as $\ell(a,y)=\mathbf{1}\{a\neq y\}$, with answer correctness
determined by the task-specific evaluation rule. The
\emph{accepted-error loss} of a fixed cascade policy is
\begin{equation}
Z_{\mathrm{acc}}(x,y)
=
\mathbf{1}\{x\text{ is accepted by a cascade stage}\}
\,\ell\!\left(a_{\mathrm{out}}(x),y\right).
\label{eq:accepted_error_definition}
\end{equation}
This loss equals one when a stage accepts an incorrect answer and
zero otherwise. Its expectation,
$\mathbb{E}[Z_{\mathrm{acc}}(X,Y)]$, is the system-level
accepted-error risk. Queries rejected by all stages are handled
by the pre-specified output rule and contribute zero to this loss.

\vspace{0.5\baselineskip}
\noindent\textbf{Stage-wise accepted-error risk contribution.}
The \emph{accepted-error risk contribution} of stage $k$ is the
probability that a random query is accepted at that stage and
answered incorrectly. The corresponding \emph{stage contribution loss}
is
\begin{equation}
Z_k(x,y)
=
\mathbf{1}\{\mathrm{reach}_k(x)\}
\,\mathbf{1}\{c_k(x)\geq t_k\}
\,\ell\!\left(a_k(x),y\right).
\label{eq:stage_contribution}
\end{equation}
The three factors indicate that the query reaches the stage,
meets its acceptance criterion, and receives an incorrect answer,
respectively. The expectation $\mathbb{E}[Z_k(X,Y)]$ is the
accepted-error risk contribution of stage $k$. This quantity is
also defined over the full query population. 

For example, suppose
a stage accepts 10 of 100 queries and answers two of them
incorrectly. Its empirical error rate among accepted answers is
20\%, whereas its empirical contribution to system-level
accepted-error risk is 2\%. Because each query is accepted by
at most one stage, the stage-wise risk contributions sum to the
system-level accepted-error risk.

\vspace{0.5\baselineskip}
\noindent\textbf{Final accuracy and inference cost.}
The \emph{final accuracy} is
$\mathbb{P}\{a_{\mathrm{out}}(X)=Y\}$.
It evaluates all delivered answers, including those returned by
the pre-specified output rule after rejection by all stages.

Let $d_k>0$ denote the normalized cost of invoking stage $k$.
The \emph{cumulative inference cost} for query $x$ is
\begin{equation}
\mathcal{C}(x)
=
\sum_{k=1}^{K}
d_k\,\mathbf{1}\{\mathrm{reach}_k(x)\}.
\label{eq:cascade_cost}
\end{equation}
This cost includes every model actually invoked for the query.
The \emph{expected inference cost} is
$\mathbb{E}[\mathcal{C}(X)]$.

\vspace{0.5\baselineskip}
\noindent\textbf{Problem statement.}
Given the model cascade, finite labeled development data $D$,
and a global risk budget $\alpha\in(0,1)$, the
\emph{risk-controlled cascade deployment problem} is to determine
an acceptance policy and a deployment procedure. 
The development
data $D$ are used to determine the policy and evaluate its risk
statistically. Let $\Pi_D$ denote the complete deployment procedure
determined from these data, including its acceptance policy and
final-output rules. 
Let $L_{\mathrm{acc}}(\Pi_D;x,y)$ denote the
accepted-error loss of this procedure. It equals one when the
procedure accepts an incorrect answer through a cascade stage
and zero otherwise. The objective is to balance final accuracy
and expected inference cost subject to
\begin{equation}
\mathbb{E}_{D,(X,Y)}
\!\left[
L_{\mathrm{acc}}(\Pi_D;X,Y)
\right]
\leq \alpha.
\label{eq:problem_target}
\end{equation}

The expectation is taken jointly over the development data and
a future query. The constraint requires the complete procedure's
marginal expected accepted-error risk to remain within the global
budget $\alpha$. Final accuracy and inference cost are evaluated
using the answers actually delivered and the models actually
invoked by the procedure. The procedure should also provide
estimates of stage-wise risk contributions to support local
risk analysis and diagnosis.

\section{Method}
\label{sec:method}

\begin{figure}[pos=h!]
    \centering
    \includegraphics[width=0.98\linewidth]{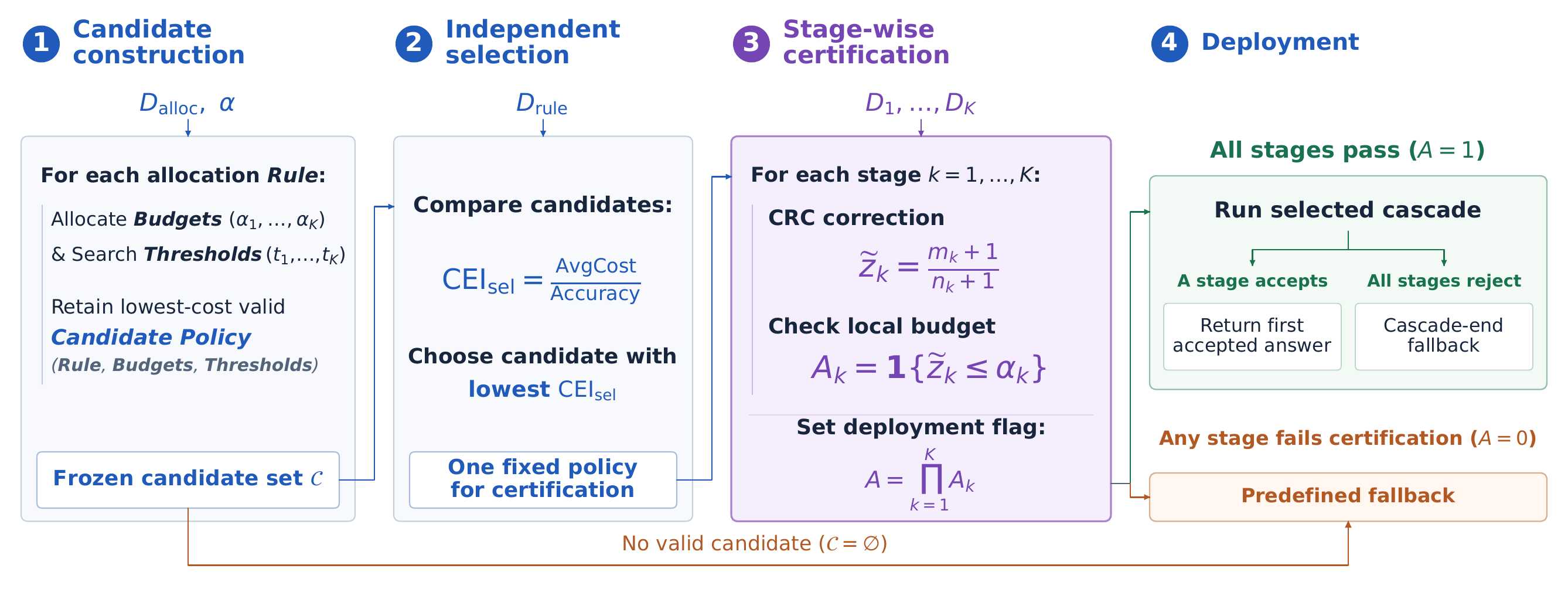}
    \caption{Overview of CompCRC. Candidate policies are fixed before independent selection and stage-wise certification. Here, $K$ denotes the number of cascade stages, $m_k$ counts incorrect answers accepted at stage $k$ on $D_k$, and $n_k=|D_k|$. The selected cascade is enabled only if all stages pass certification. An empty candidate set or certification failure triggers the predefined fallback. For an enabled cascade, cascade-end fallback returns the highest-confidence answer among the already generated outputs when all stages reject a query.}
    \label{fig:framework}
\end{figure}

We propose \emph{Compositional Conformal Risk Control} (CompCRC) to control the risk of accepting incorrect answers in multi-stage large language model (LLM) cascades by distributing a global risk budget across individual stages. Consider a predefined, fixed cascade with $K$ stages, where $K$ is a finite positive integer and $k\in\{1,\ldots,K\}$ denotes the stage index. We present the method for a general $K$-stage cascade; the experiments instantiate it with $K=3$. Fig.~\ref{fig:framework} illustrates the overall workflow, and Algorithm~\ref{alg:compcrc} summarizes its execution. CompCRC first constructs candidate configurations consisting of stage-wise risk budgets and acceptance thresholds. It then uses an independent selection set to identify a configuration that balances accuracy and inference cost, and finally evaluates each stage's risk contribution using dedicated certification data. The selected cascade is deployed only if all stages pass certification; otherwise, a predefined certification-failure fallback policy is used.

\begin{algorithm}[t]
\caption{Compositional Conformal Risk Control (CompCRC)}
\label{alg:compcrc}
\begin{algorithmic}[1]
\Require Global risk budget $\alpha$; fixed $K$-stage cascade and generation settings; budget allocation and search configurations; datasets $D_{\mathrm{alloc}},D_{\mathrm{rule}},D_1,\ldots,D_K$; calibrated Route-only baseline and upper bound on the cost ratio; predefined certification-failure fallback policy.
\Ensure Selected configuration and certification record (if available), execution status, and deployment policy $\Pi_D$.
\Statex \textbf{Candidate construction}
\State Initialize the candidate set $\mathcal C\gets\varnothing$.
\For{each budget allocation rule}
    \State Construct and screen configurations on $D_{\mathrm{alloc}}$ as described in Section~\ref{subsec:method_candidates}.
    \If{at least one feasible configuration exists}
        \State Retain the one with the lowest average invocation cost; fix its budgets and thresholds, and add it to $\mathcal C$.
    \EndIf
\EndFor
\If{$\mathcal C=\varnothing$}
    \State Set $\Pi_D$ to the predefined fallback policy.
    \State \Return ``no feasible candidate'' status and $\Pi_D$.
\EndIf
\Statex \textbf{Independent selection}
\State Evaluate $\mathrm{CEI}_{\mathrm{sel}}$ for each candidate on $D_{\mathrm{rule}}$ using Eq.~\eqref{eq:selection_cei}, and select a minimizer for certification.
\Statex \textbf{Stage-wise certification}
\For{$k=1,\ldots,K$}
    \State Run the selected cascade on \emph{all} queries in $D_k$ and compute $m_k$, $\widehat z_k$, $\widetilde z_k$, and $A_k$ using Eqs.~\eqref{eq:stage_empirical}--\eqref{eq:certification_flags}.
\EndFor
\State Compute the deployment indicator $A\gets\prod_{k=1}^{K} A_k$.
\Statex \textbf{Deployment}
\If{$A=1$}
    \State Set $\Pi_D$ to the selected cascade: return the first stage answer meeting its fixed acceptance threshold; if none is accepted, return the highest-confidence answer among those generated.
\Else
    \State Mark the selected operating point as not certified (NC) and set $\Pi_D$ to the predefined certification-failure fallback policy.
\EndIf
\State \Return the selected configuration, the stage-wise records $(\alpha_k,\widehat z_k,\widetilde z_k,A_k)$, $A$, and $\Pi_D$.
\end{algorithmic}
\end{algorithm}

\subsection{Risk decomposition and local budgets}
\label{subsec:method_risk}

CompCRC decomposes the system-level accepted-error risk across stages by exploiting the mutual exclusivity of stage acceptance events. Under the definitions in Section~3, each query can be accepted by at most one stage. Consequently, the system-level accepted-error loss satisfies $Z_{\mathrm{acc}}(x,y)=\sum_{k=1}^{K}Z_k(x,y)$. Taking expectations gives
\begin{equation}
\label{eq:risk_decomposition}
\mathbb{E}\!\left[Z_{\mathrm{acc}}(X,Y)\right]
=\sum_{k=1}^{K}\mathbb{E}\!\left[Z_k(X,Y)\right].
\end{equation}
Equation~\eqref{eq:risk_decomposition} establishes that the system-level accepted-error risk equals the sum of the stage-wise risk contributions, providing the basis for local budget allocation. CompCRC assigns a local risk budget $\alpha_k$ to stage $k$ to constrain its contribution to the accepted-error risk over \emph{all} queries, subject to
\begin{equation}
\label{eq:local_budgets}
\alpha_k\geq 0,\qquad\sum_{k=1}^{K}\alpha_k=\alpha,
\end{equation}
where $\alpha$ is the global risk budget. Thus, if the accepted-error risk contribution of every stage is bounded by its allocated budget, Eq.~\eqref{eq:risk_decomposition} implies that the system-level accepted-error risk does not exceed $\alpha$.

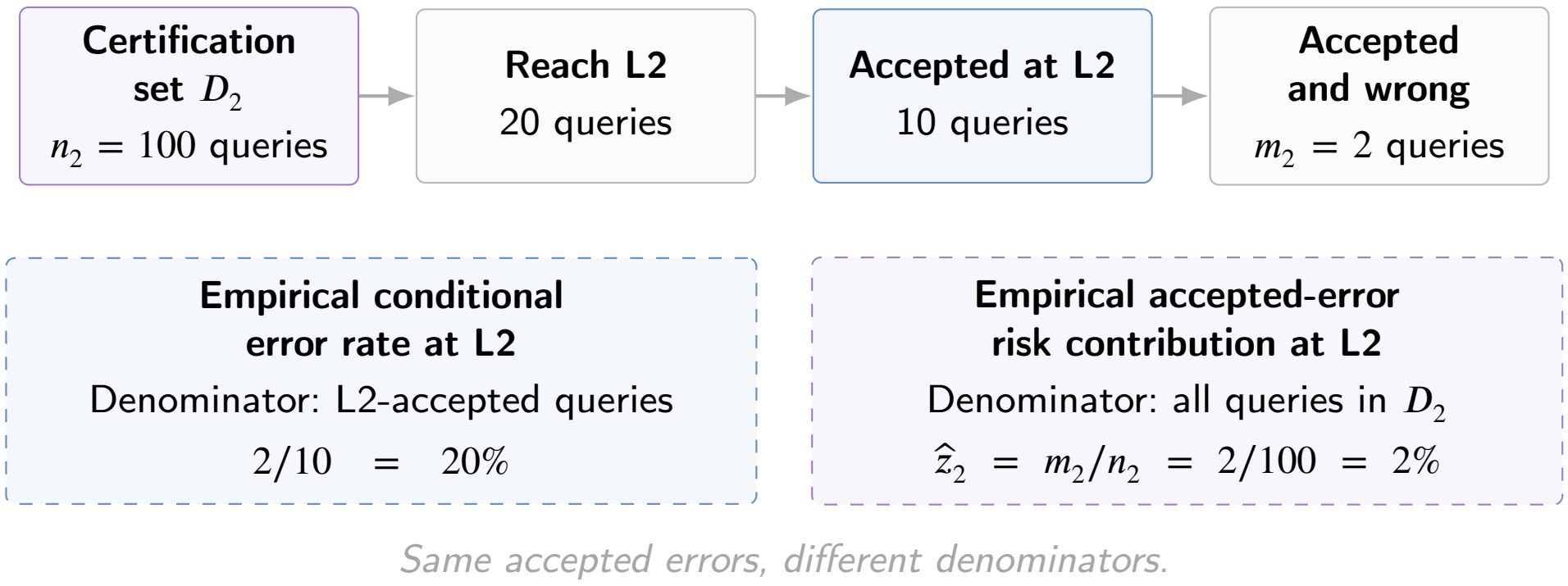
\begin{figure*}[pos=h!]
\centering
\definecolor{figtwoblue}{HTML}{205BBB}
\definecolor{figtwopurple}{HTML}{7546B3}
\begin{tikzpicture}[
    flowbox/.style={
        draw=gray!55,
        fill=gray!3,
        rounded corners=2pt,
        minimum height=1.40cm,
        text width=2.45cm,
        align=center,
        inner sep=4pt,
        font=\small
    },
    flowarrow/.style={
        -{Latex[length=2.2mm]},
        thick,
        draw=gray!65
    },
    metricbox/.style={
        dashed,
        rounded corners=2pt,
        minimum height=1.85cm,
        text width=5.70cm,
        align=center,
        inner sep=5pt,
        font=\small
    }
]

\node[flowbox,draw=figtwopurple!70,fill=figtwopurple!5] (all) at (-4.8,0)
    {\textbf{Certification set $D_2$}\\[2pt]
     $n_2=100$ queries};
\node[flowbox] (reach) at (-1.6,0)
    {\textbf{Reach L2}\\[2pt]
     20 queries};
\node[flowbox,draw=figtwoblue!70,fill=figtwoblue!5] (accept) at (1.6,0)
    {\textbf{Accepted at L2}\\[2pt]
     10 queries};
\node[flowbox] (error) at (4.8,0)
    {\textbf{Accepted and wrong}\\[2pt]
     $m_2=2$ queries};

\draw[flowarrow] (all.east) -- (reach.west);
\draw[flowarrow] (reach.east) -- (accept.west);
\draw[flowarrow] (accept.east) -- (error.west);

\node[metricbox,draw=figtwoblue!70,fill=figtwoblue!5]
    (conditional) at (-3.25,-2.30)
    {\textbf{Empirical conditional}\\
     \textbf{error rate at L2}\\[2pt]
     Denominator: L2-accepted queries\\[3pt]
     $2/10=20\%$};

\node[metricbox,draw=figtwopurple!70,fill=figtwopurple!5]
    (contribution) at (3.25,-2.30)
    {\textbf{Empirical accepted-error}\\
     \textbf{risk contribution at L2}\\[2pt]
     Denominator: all queries in $D_2$\\[3pt]
     $\widehat z_2=m_2/n_2=2/100=2\%$};

\node[
    font=\small\itshape,
    text=gray!70,
    anchor=north
] at (0,-3.50)
    {Same accepted errors, different denominators.};
    
\end{tikzpicture}
\caption{Illustration of the empirical conditional error rate and
accepted-error risk contribution at L2.}
\label{fig:risk_example}
\end{figure*}

Fig.~\ref{fig:risk_example} illustrates the distinction between a stage-conditional error rate and a stage's contribution to system-level risk, using stage L2 in a three-stage cascade. Suppose that L2 accepts 10 of the 100 queries in its certification set $D_2$, and two of those accepted answers are incorrect. Its empirical error rate conditional on acceptance is $2/10=20\%$, whereas its empirical contribution to system-level accepted-error risk is $2/100=2\%$. Normalizing by the total number of queries, rather than the number accepted at that stage, puts all stage-wise contributions on the same scale and is consistent with the additive decomposition in Eq.~\eqref{eq:risk_decomposition} and the budget allocation in Eq.~\eqref{eq:local_budgets}.

\subsection{Candidate construction and selection}
\label{subsec:method_candidates}

CompCRC constructs candidate configurations under different risk-budget allocation rules, determining each configuration's local risk budgets and acceptance thresholds. An independent dataset is then used to compare the accuracy and inference cost of these fixed candidates and select one for certification.

\paragraph{Data partitioning.}
The development data are partitioned according to the three steps of candidate construction, independent selection, and stage-wise certification. The construction set $D_{\mathrm{alloc}}$ is used to determine local budgets and acceptance thresholds. The selection set $D_{\mathrm{rule}}$ is used to compare configurations fixed after construction. Finally, the stage-specific certification sets $D_1,\ldots,D_K$ are used to evaluate the selected configuration's risk contributions. These mutually disjoint subsets together constitute the development set $D$ defined in Section~3:
\begin{equation}
\label{eq:data_partition}
D=D_{\mathrm{alloc}}\cup D_{\mathrm{rule}}\cup\bigcup_{k=1}^{K}D_k.
\end{equation}
The construction set $D_{\mathrm{alloc}}$ and the $K$ certification sets $D_1,\ldots,D_K$ are obtained by partitioning the CompCRC calibration pool $D_{\mathrm{cal}}$, while $D_{\mathrm{rule}}$ is sampled separately from the same underlying dataset. A separate calibration set $D_{\mathrm{route}}$ is used to determine the parameters of the Route-only baseline. The final test set is disjoint from all these subsets and is reserved for evaluating risk, accuracy, and inference cost. Table~\ref{tab:data_partitions} summarizes their roles.

\begin{table}[t]
\centering
\caption{Data partitions and their roles in CompCRC.}
\label{tab:data_partitions}
\small
\begin{tabularx}{\linewidth}{@{}l l X@{}}
\toprule
\textbf{Partition} & \textbf{Purpose} & \textbf{Usage} \\
\midrule
$D_{\mathrm{alloc}}$ & Candidate construction & Determine and fix local budgets and acceptance thresholds before independent selection. \\
$D_{\mathrm{rule}}$ & Candidate selection & Compare the accuracy and average invocation cost of fixed candidates. \\
$D_1,\ldots,D_K$ & Stage-wise certification & Independently assess the selected configuration's risk contribution at each stage. \\
$D_{\mathrm{route}}$ & Baseline calibration & Determine the routing parameters of the Route-only baseline. \\
Test set & Final evaluation & Report risk, accuracy, and inference cost. \\
\bottomrule
\end{tabularx}
\end{table}

\paragraph{Budget allocation rules.}
CompCRC allocates local risk budgets using model invocation costs, query arrival proportions, and empirical error rates. Let $p_k$ denote the proportion of queries reaching stage $k$ on the construction set, and let $e_k$ denote the empirical error rate among answers accepted at that stage. Both quantities are estimated by executing the cascade under its current thresholds. For a stage with model invocation cost $d_k$, define the cost factor $u_k=\eta_k/d_k$, where $\eta_k>0$ is a stage-specific adjustment coefficient fixed before candidate construction. Let $\beta=(\beta_c,\beta_f,\beta_e)\in\{0,1\}^3$ indicate whether the cost, flow, and error-rate factors are enabled, respectively. The allocation weight for stage $k$ is
\begin{equation}
\label{eq:allocation_weights}
w_k(\beta)=u_k^{\beta_c}p_k^{-\beta_f/2}(1+e_k)^{\beta_e}.
\end{equation}
Normalizing these weights yields the local risk budget
\begin{equation}
\label{eq:normalized_budget}
\alpha_k=\alpha\frac{w_k(\beta)}{\sum_{j=1}^{K}w_j(\beta)},
\end{equation}
which ensures that the local budgets sum to the global budget $\alpha$.

The eight possible values of $\beta$ give the allocation rules listed in Table~\ref{tab:allocation_rules}. Holding the other factors fixed, the cost factor allocates more budget to less expensive stages; the flow factor favors stages reached by smaller proportions of queries; and the error-rate factor increases the allocation to stages with higher empirical error rates. The term $1+e_k$ ensures that the error-rate factor remains positive. Their combinations allow CompCRC to account for different stage characteristics when constructing candidate allocations.

\begin{table}[t]
\centering
\caption{Candidate risk-budget allocation rules.}
\label{tab:allocation_rules}
\begin{tabular}{@{}ll@{}}
\toprule
\textbf{Rule} & \textbf{Stage weight $w_k$} \\
\midrule
Uniform & $1$ \\
Cost & $u_k$ \\
Flow & $p_k^{-1/2}$ \\
Error & $1+e_k$ \\
Cost--flow & $u_kp_k^{-1/2}$ \\
Cost--error & $u_k(1+e_k)$ \\
Flow--error & $p_k^{-1/2}(1+e_k)$ \\
Full & $u_kp_k^{-1/2}(1+e_k)$ \\
\bottomrule
\end{tabular}
\end{table}

\paragraph{Candidate construction.}
Each candidate configuration consists of an allocation rule, local risk budgets $(\alpha_1,\ldots,\alpha_K)$, and acceptance thresholds $(t_1,\ldots,t_K)$. Its construction and screening correspond to the candidate-construction step of Algorithm~\ref{alg:compcrc}. For each allocation rule, CompCRC alternates between estimating execution statistics, updating budgets, and searching acceptance thresholds on $D_{\mathrm{alloc}}$. Execution statistics estimate the arrival proportions and empirical error rates under the current thresholds. Threshold search enumerates combinations of thresholds across the $K$ stages and simulates cascade execution to assess their performance. To handle zero arrival proportions in weight computation, a predefined small positive lower bound is used. Configurations for which any stage has no accepted examples are discarded.

Candidate screening jointly considers risk, accuracy, and invocation cost. Accuracy and cost are compared with those of the Route-only baseline evaluated on the same construction set. This baseline routes queries according to confidence thresholds, with routing parameters calibrated on $D_{\mathrm{route}}$. A candidate is feasible only if (i) every stage's empirical accepted-error risk contribution does not exceed its local budget and the sum of these empirical contributions does not exceed $\alpha$; (ii) its final accuracy is no lower than that of Route-only; and (iii) its ratio of average invocation cost to the Route-only average invocation cost does not exceed a predefined upper bound. In our experiments, this upper bound is set to $1.5$, allowing up to a 50\% increase in average invocation cost relative to the baseline to accommodate more conservative acceptance policies.

For each allocation rule, CompCRC retains the feasible configuration with the lowest average invocation cost and adds it to the candidate set $\mathcal C$. Rules that produce no feasible configurations are excluded from subsequent selection. The local budgets and acceptance thresholds of all retained candidates are fixed at the end of construction and remain unchanged during selection and certification. If $\mathcal C=\varnothing$, the procedure returns a ``no feasible candidate'' status and directly applies the same predefined fallback policy used after certification failure, without proceeding to selection or certification.

\paragraph{Independent candidate selection.}
CompCRC evaluates the fixed candidates on $D_{\mathrm{rule}}$ and selects one for certification based on accuracy and invocation cost. Let $\mathrm{AvgCost}_{D_{\mathrm{rule}}}$ and $\mathrm{Accuracy}_{D_{\mathrm{rule}}}$ denote a candidate's average invocation cost and final accuracy on this selection set, respectively. Define the selection cost-efficiency index as
\begin{equation}
\label{eq:selection_cei}
\mathrm{CEI}_{\mathrm{sel}}=
\frac{\mathrm{AvgCost}_{D_{\mathrm{rule}}}}{\mathrm{Accuracy}_{D_{\mathrm{rule}}}}.
\end{equation}
This index measures average invocation cost per unit of accuracy, with smaller values indicating better cost efficiency. It is defined as $+\infty$ when accuracy is zero; ties for the minimum are resolved using a predefined rule. CompCRC selects a candidate minimizing $\mathrm{CEI}_{\mathrm{sel}}$. Together, the global risk budget $\alpha$, its selected local budgets, and its acceptance thresholds define an operating point that is subsequently subjected to stage-wise certification.

\subsection{Stage-wise certification and deployment}
\label{subsec:method_certification}

CompCRC certifies the selected, fixed configuration stage by stage before deciding whether to deploy it. Stage-specific risk statistics are also reported to support diagnosis. Two fallback mechanisms are distinguished: a system-level fallback when certification fails and a terminal fallback when a deployed cascade accepts no stage answer for a particular query.

\paragraph{Stage-wise certification.}
For each stage $k$, let $D_k=\{(X_{ki},Y_{ki})\}_{i=1}^{n_k}$ denote its dedicated certification set, with $n_k\geq 1$. The selected cascade is executed on \emph{all} queries in $D_k$ to evaluate the stage-contribution loss $Z_k$. This loss equals one only when the answer is accepted at stage $k$ and is incorrect, and equals zero otherwise. Summing these losses gives the number of incorrectly accepted answers $m_k$. Dividing by the \emph{full} certification-set size yields the empirical risk contribution $\widehat z_k$:
\begin{equation}
\label{eq:stage_empirical}
m_k=\sum_{i=1}^{n_k}Z_k(X_{ki},Y_{ki}),\qquad
\widehat z_k=\frac{m_k}{n_k}.
\end{equation}

Because the stage loss is bounded by one, CompCRC applies a finite-sample correction used in conformal risk control (Angelopoulos et al., 2024). This correction is equivalent to appending one worst-case observation with loss one to the certification sample and dividing by $n_k+1$. The resulting corrected certification statistic is
\begin{equation}
\label{eq:corrected_statistic}
\widetilde z_k=
\frac{n_k}{n_k+1}\widehat z_k+\frac{1}{n_k+1}
=\frac{m_k+1}{n_k+1}.
\end{equation}

Stage $k$ passes certification if $\widetilde z_k\leq\alpha_k$ and fails otherwise. Let $A_k$ denote its pass indicator. The selected cascade is deployed only if every stage passes; hence, the system deployment indicator is the product of the stage indicators:
\begin{equation}
\label{eq:certification_flags}
A_k=\mathbf{1}\{\widetilde z_k\leq\alpha_k\},\qquad
A=\prod_{k=1}^{K}A_k.
\end{equation}

\paragraph{Deployment and fallback.}
Once the selected operating point has been certified, CompCRC determines the deployment policy according to $A$. If $A=1$, the system deploys the selected cascade with its fixed acceptance thresholds. If $A=0$, the operating point is marked \emph{not certified} (NC), and the selected cascade is replaced by the predefined certification-failure fallback policy.

For a deployed cascade, execution stops as soon as a stage satisfies its acceptance condition, and the corresponding answer is returned. If no stage accepts the query, the cascade applies its terminal fallback policy, returning one of the answers already generated. Here, the answer with the highest confidence is selected, without any additional model invocation.

Outputs produced by either fallback mechanism are included in final accuracy calculations, while inference cost is measured from the model invocations actually performed. As these fallback outputs are not designated as stage-accepted answers, they do not contribute to the stage-wise accepted-error loss under the definitions in Section~3. When no feasible candidate is found, the predefined fallback is used directly, likewise yielding zero accepted-error loss. The accepted-error loss of the complete deployment procedure is therefore
\begin{equation}
\label{eq:deployment_loss}
L_{\mathrm{acc}}(\Pi_D;x,y)=
\begin{cases}
A\,Z_{\mathrm{acc}}(x,y), & \mathcal C\neq\varnothing,\\[2pt]
0, & \mathcal C=\varnothing.
\end{cases}
\end{equation}
The finite-sample risk guarantee for this deployment procedure is presented in Section~5.

\paragraph{Stage-wise risk diagnostics.}
For each stage, CompCRC reports its local budget $\alpha_k$, empirical risk contribution $\widehat z_k$, corrected certification statistic $\widetilde z_k$, and pass indicator $A_k$. Diagnostics compare the corrected statistic with the local budget: stage $k$ fails its local check when $\widetilde z_k>\alpha_k$. Reporting these results separately identifies which stages prevent the selected operating point from being certified and shows how their empirical risk contributions and corrected statistics compare with the corresponding local budgets.

\Needspace{12\baselineskip}

\section{Theoretical Analysis}
\label{sec:theoretical_analysis}

This section establishes a finite-sample guarantee for the accepted-error risk of the complete CompCRC deployment procedure. Candidate construction and selection determine the configuration to be certified, together with its local budgets and acceptance thresholds. The stage-wise certification outcomes jointly determine whether the selected cascade is deployed. We first establish the validity of stage-wise certification and then use the additive risk decomposition to derive a system-level guarantee that meets the risk objective in Section~3.

\subsection{Certification conditions and stage-wise guarantee}
\label{subsec:theory_stage}

Let $\mathcal F_0$ denote all information used for candidate construction, independent selection, and system configuration before certification. This includes the relevant data, the selected configuration, local budgets, acceptance thresholds, and model generation settings. Conditional on $\mathcal F_0$, the candidate set $\mathcal C$ is fixed. When $\mathcal C\neq\varnothing$, the selected configuration and its parameters are also fixed, and the local budgets satisfy Eq.~\eqref{eq:local_budgets}. The certification sample sizes $n_k\geq1$ are specified before certification. Thus, conditional on $\mathcal F_0$, the analysis concerns the selected configuration with fixed parameters.

When $\mathcal C\neq\varnothing$, assume that, for each stage $k$, the $n_k$ stage-contribution losses of the selected configuration on $D_k$ and its stage-contribution loss on a future query are jointly exchangeable conditional on $\mathcal F_0$, with all losses taking values in $[0,1]$. For stochastic generation, the losses incorporate randomness in model outputs, and certification and online inference use the same generation mechanism. Conditional exchangeability provides the statistical basis for stage-wise certification. Candidate construction and selection are completed before accessing the certification data, and the selected budgets and thresholds remain fixed throughout certification.

\begin{lemma}[Validity of stage-wise certification]
\label{lem:stage_certification}
When $\mathcal C\neq\varnothing$, under the conditions above, the stage pass indicator $A_k$ defined in Eq.~\eqref{eq:certification_flags} satisfies
\begin{equation}
\label{eq:stage_certification_bound}
\mathbb E\!\left[A_k Z_k(X,Y)\mid\mathcal F_0\right]
\leq\alpha_k,
\qquad k=1,\ldots,K.
\end{equation}
\end{lemma}

\begin{proof}
Fix a stage $k$ and write $n=n_k$ and $b=\alpha_k$. Let $W_i=Z_k(X_{ki},Y_{ki})$, $i=1,\ldots,n$, denote the certification losses, and let $W_{n+1}=Z_k(X,Y)$ denote the loss on a future query. These $n+1$ losses are exchangeable conditional on $\mathcal F_0$. Write $S=\sum_{i=1}^{n+1}W_i$ and define the certification indicator obtained by leaving out observation $i$ as
\begin{equation}
\label{eq:leave_one_out_certification}
B_{-i}
=
\mathbf 1\!\left\{
\frac{S-W_i+1}{n+1}\leq b
\right\},
\qquad i=1,\ldots,n+1.
\end{equation}
Leaving out the future query gives $B_{-(n+1)}=A_k$. Since $W_i\leq1$, replacing the omitted loss $W_i$ with the worst-case loss $1$ gives $S-W_i+1\geq S$. Hence, $B_{-i}=1$ implies $S/(n+1)\leq b$. If at least one $B_{-i}=1$, nonnegativity of the losses gives $\sum_iW_iB_{-i}/(n+1)\leq S/(n+1)\leq b$; if all $B_{-i}=0$, this average equals zero. Conditional exchangeability ensures that the terms $W_iB_{-i}$ have the same conditional expectation. Therefore,
\begin{equation}
\label{eq:stage_certification_proof}
\begin{aligned}
\mathbb E\!\left[A_k Z_k(X,Y)\mid\mathcal F_0\right]
&=
\mathbb E\!\left[
\frac{1}{n+1}\sum_{i=1}^{n+1}W_iB_{-i}
\,\middle|\,\mathcal F_0
\right]\\
&\leq b=\alpha_k.
\end{aligned}
\end{equation}
\end{proof}

The finite-sample correction in
Eq.~\eqref{eq:corrected_statistic} thus yields the risk
guarantee in Eq.~\eqref{eq:stage_certification_bound}
for stage-wise certification of the selected configuration.
If $\alpha_k<1/(n_k+1)$, even zero observed errors
yield $A_k=0$, so the system uses the predefined
certification-failure fallback.

\subsection{System-level risk guarantee}
\label{subsec:theory_system}

CompCRC deploys the selected cascade only when all stages pass certification. The system deployment indicator therefore satisfies $A\leq A_k$ for $k=1,\ldots,K$. Combining this relation with the additive decomposition of stage-wise risks extends Lemma~\ref{lem:stage_certification} to the complete deployment procedure.

\begin{theorem}[Risk control for the complete deployment procedure]
\label{thm:deployment_risk}
For a prespecified global risk budget $\alpha\in(0,1)$, under the conditions in Section~\ref{subsec:theory_stage}, the CompCRC deployment policy $\Pi_D$ defined in Section~\ref{sec:method} satisfies
\begin{equation}
\label{eq:deployment_risk_bound}
\mathbb E\!\left[
L_{\mathrm{acc}}(\Pi_D;X,Y)
\mid\mathcal F_0
\right]
\leq\alpha.
\end{equation}
The conditional expectation averages over the certification data, the future query, and the corresponding randomness in model generation.
\end{theorem}

\begin{proof}
When $\mathcal C=\varnothing$, the system directly applies the predefined fallback policy. Equation~\eqref{eq:deployment_loss} gives zero accepted-error loss, establishing the claim in this case. When $\mathcal C\neq\varnothing$, the identity $A=\prod_{k=1}^{K}A_k$ implies $A\leq A_k$. Using the loss decomposition $Z_{\mathrm{acc}}=\sum_{k=1}^{K}Z_k$ from Section~\ref{subsec:method_risk}, the nonnegativity of $Z_k$, and Lemma~\ref{lem:stage_certification}, we obtain
\begin{equation}
\label{eq:deployment_risk_composition}
\begin{aligned}
\mathbb E\!\left[
A Z_{\mathrm{acc}}(X,Y)\mid\mathcal F_0
\right]
&=
\sum_{k=1}^{K}
\mathbb E\!\left[A Z_k(X,Y)\mid\mathcal F_0\right]\\
&\leq
\sum_{k=1}^{K}
\mathbb E\!\left[A_k Z_k(X,Y)\mid\mathcal F_0\right]\\
&\leq
\sum_{k=1}^{K}\alpha_k
=\alpha.
\end{aligned}
\end{equation}
Together with Eq.~\eqref{eq:deployment_loss}, this establishes Eq.~\eqref{eq:deployment_risk_bound}. Taking expectations over the information $\mathcal F_0$ available before certification and applying the law of total expectation yields the unconditional risk guarantee required by Eq.~(4) in Section~3.
\end{proof}

Theorem~\ref{thm:deployment_risk} connects candidate selection, stage-wise certification, and deployment. The construction data determine and fix candidate configurations, the selection data identify the configuration to be certified, and the certification data determine the stage pass indicators. Deploying the cascade only after all stages pass combines the stage-wise guarantees into a system-level guarantee. This composition relies on nonnegative losses, their additive decomposition, and $A\leq A_k$. It also applies when confidence scores or certification outcomes are correlated across stages.

Theorem~\ref{thm:deployment_risk} provides a marginal accepted-error risk guarantee for the complete deployment procedure, including certification and fallback. Conditional on the information $\mathcal F_0$ available before certification, the guarantee averages over the certification data, the future query, and randomness in model generation. The accepted-error loss follows the definition in Section~3 and counts errors in stage-accepted answers. Under this definition, fallback outputs incur zero accepted-error loss; their predictive performance and computational overhead are included in final accuracy and actual invocation cost, respectively. The experiments evaluate CompCRC in terms of risk, accuracy, and inference cost using these definitions.

\section{Experiments}
\label{sec:experiments}

The experiments address four research questions, corresponding to
Sections~\ref{subsec:main_results}--\ref{subsec:budget_feasibility}.

\smallskip
\noindent\textbf{RQ1: Overall performance comparison.}
How does CompCRC compare with the baselines in observed risk, accuracy,
and cost?

\smallskip
\noindent\textbf{RQ2: Stage-wise certification and diagnosis.}
How do stage-wise checks assess budget utilization and identify the
stages that fail certification?

\smallskip
\noindent\textbf{RQ3: Effect of risk-budget allocation.}
How does cost-based allocation compare with uniform allocation in
performance and stage-wise budget utilization?

\smallskip
\noindent\textbf{RQ4: Sensitivity to the global risk budget.}
How does changing the global risk budget affect observed risk, cost,
and certification feasibility?

\subsection{Experimental Setup}
\label{subsec:experimental_setup}

\paragraph{Datasets and Inference Settings}
We evaluate on MMLU \citep{Hendrycks2021MMLU}, GSM8K
\citep{Cobbe2021GSM8K}, and HellaSwag \citep{Zellers2019HellaSwag}.
The three-stage cascade consists of Llama-3.2-3B-Instruct (L1),
GLM-4-9B-0414 (L2), and Qwen3-14B (L3), ordered by increasing
reference API cost. MMLU and HellaSwag use four-option prompts
with a maximum generation length of 120 tokens. GSM8K uses a
maximum of 192 tokens and requires the final answer in the format
\texttt{\#\#\#\# <number>}. The sampling temperature is 0.1 for
all tasks. Answers are evaluated by extracting A/B/C/D for the
multiple-choice tasks and the value following the final
\texttt{\#\#\#\#} marker for GSM8K.

For MMLU and HellaSwag, confidence is the geometric mean of
the generated-token probabilities. GSM8K uses a conservative
confidence score for the final-answer tokens, with confidence
set to zero when parsing fails or generation is incomplete.
Inference uses the frozen stage-specific acceptance thresholds
and the cascade-end fallback described in
Section~\ref{sec:method}. The fallback requires no additional
model call.

\paragraph{Data Splits and Evaluation Protocol}
Each dataset has mutually disjoint routing-calibration,
CompCRC-calibration, rule-selection, and test sets containing
1,500, 1,275, 1,000, and 300 samples, respectively.
The CompCRC calibration pool is divided into 255 samples for
candidate construction and 318, 318, and 384 samples for the
stage-specific certification sets $D_1$, $D_2$, and $D_3$.
Construction, selection, and certification samples are randomly
shuffled before splitting. The test set is sampled independently
and used only for final evaluation. All methods are evaluated
on the same test set for each dataset.

\paragraph{Comparison Methods}
We evaluate CompCRC using the protocol in
Section~\ref{sec:method}, together with our Route-Gated CompCRC
variant, which adds routing-threshold constraints to examine
their effect on performance and cost.

We compare these two methods with six baselines.
L1 only, L2 only, and L3 only serve as single-model
performance--cost references.
Route-only, a confidence-based routing baseline, learns
three-stage routing thresholds using $D_{\mathrm{route}}$.
FrugalGPT-style, an empirical accuracy--cost baseline, uses
a fixed model pool and a pre-specified aggregation configuration.
Global-CRC selects and freezes a global threshold on an
independent design set, then certifies the resulting policy
on a disjoint certification set. Certification failure
triggers its pre-specified fallback.
Formal CRC guarantees apply to CompCRC (stage-wise) and
Global-CRC (aggregate); Route-Gated CompCRC and the other
baselines are evaluated without such guarantees.

\paragraph{Evaluation Metrics}
We report observed risk following
Section~\ref{sec:problem_formulation}, together with accuracy,
cost ratio, and Acc./Cost. For single-model baselines,
every output is treated as accepted.
Cost normalization uses the mean Route-only cost on the
same dataset as the reference value of 1.0.
The invocation weights approximate input-token price plus
three times output-token price. Normalizing these weights
to L1 gives
$\boldsymbol d=(d_1,d_2,d_3)=(1.0000,1.6667,4.7619)$,
with each visited stage contributing its weight to the
query's invocation cost.

\paragraph{Certification Criteria and Risk Budgets}
CompCRC uses the stage-wise certification criteria in
Section~\ref{sec:method}, while Global-CRC uses aggregate
certification. Certification outcomes are reported separately
from test performance. The tables indicate the formal
guarantee and CRC status, with N/A for methods without CRC
certification. We evaluate the methods separately at each
budget in a common discrete grid and report $\alpha=0.45$
as the common comparison point for the main results, at
which Global-CRC and CompCRC pass their respective
certification procedures on all three datasets. The guarantee
in Theorem~\ref{thm:deployment_risk} applies to the CompCRC
deployment procedure at each global risk budget fixed before
certification, but does not cover post hoc budget selection
based on certification outcomes. The budget sweeps describe
the resulting fixed-budget operating points and are reported in
Section~\ref{subsec:budget_feasibility} and
Appendix~\ref{app:cross_dataset_sweeps}.

\subsection{Overall Performance Comparison}
\label{subsec:main_results}

To answer RQ1, we compare CompCRC and its Route-Gated variant
with six baselines on MMLU, GSM8K, and HellaSwag.
Table~\ref{tab:main_results} reports their performance on the
same independent test sets.
At the reported risk budget $\alpha=0.45$, both Global-CRC
and CompCRC pass their respective certification procedures.
Risk, accuracy, and cost are the primary performance
measures, while Acc./Cost provides a supplementary
efficiency measure.
Lower Risk and Cost and higher accuracy and Acc./Cost
are preferred.
The formal-guarantee and CRC-status columns indicate the
guarantee type and certification outcome, respectively.

\begin{table}[pos=h!]
\caption{Overall performance comparison at $\alpha=0.45$.}
\label{tab:main_results}
\centering
\begingroup
\scriptsize
\setlength{\tabcolsep}{2.5pt}
\renewcommand{\arraystretch}{1.10}
\begin{tabular*}{\linewidth}{
@{\extracolsep{\fill}}l l c r r r r@{}}
\toprule
\textbf{Method} &
\textbf{Formal guarantee} &
\textbf{CRC status} &
Risk$\downarrow$ &
Acc.$\uparrow$ &
Cost$\downarrow$ &
Acc./Cost$\uparrow$ \\
\midrule
\multicolumn{7}{l}{\textbf{MMLU}} \\
\midrule
L1 only
& None & N/A
& 0.2533 & 0.7467 & 0.6587 & 1.1336 \\
L2 only
& None & N/A
& 0.1500 & 0.8500 & 1.0978 & 0.7743 \\
L3 only
& None & N/A
& 0.1167 & 0.8833 & 3.1364 & 0.2816 \\
Route-only
& None & N/A
& 0.1733 & 0.8200 & 1.0000 & 0.8200 \\
FrugalGPT-style
& None & N/A
& 0.2067 & 0.7933 & 0.8636 & 0.9186 \\
Global-CRC
& Aggregate CRC & Pass
& 0.1800 & 0.8200 & 1.2039 & 0.6811 \\
\hdashline
CompCRC (Ours)
& Stage-wise CRC & Pass
& 0.1667 & 0.8333 & 1.2854 & 0.6483 \\
Route-Gated CompCRC (Ours)
& None & N/A
& 0.0933 & 0.8667 & 1.5604 & 0.5554 \\
\midrule
\multicolumn{7}{l}{\textbf{GSM8K}} \\
\midrule
L1 only
& None & N/A
& 0.4533 & 0.5467 & 0.6300 & 0.8678 \\
L2 only
& None & N/A
& 0.1333 & 0.8667 & 1.0500 & 0.8254 \\
L3 only
& None & N/A
& 0.0833 & 0.9167 & 3.0000 & 0.3056 \\
Route-only
& None & N/A
& 0.3000 & 0.6933 & 1.0000 & 0.6933 \\
FrugalGPT-style
& None & N/A
& 0.2900 & 0.7100 & 1.8900 & 0.3757 \\
Global-CRC
& Aggregate CRC & Pass
& 0.2533 & 0.7467 & 1.1190 & 0.6673 \\
\hdashline
CompCRC (Ours)
& Stage-wise CRC & Pass
& 0.2367 & 0.7533 & 1.1105 & 0.6783 \\
Route-Gated CompCRC (Ours)
& None & N/A
& 0.1767 & 0.8100 & 1.4205 & 0.5702 \\
\midrule
\multicolumn{7}{l}{\textbf{HellaSwag}} \\
\midrule
L1 only
& None & N/A
& 0.5833 & 0.4167 & 0.6415 & 0.6496 \\
L2 only
& None & N/A
& 0.3067 & 0.6933 & 1.0692 & 0.6484 \\
L3 only
& None & N/A
& 0.2500 & 0.7500 & 3.0550 & 0.2455 \\
Route-only
& None & N/A
& 0.3667 & 0.6333 & 1.0000 & 0.6333 \\
FrugalGPT-style
& None & N/A
& 0.3700 & 0.6300 & 2.0056 & 0.3141 \\
Global-CRC
& Aggregate CRC & Pass
& 0.3700 & 0.6300 & 1.5896 & 0.3963 \\
\hdashline
CompCRC (Ours)
& Stage-wise CRC & Pass
& 0.3500 & 0.6500 & 1.1069 & 0.5872 \\
Route-Gated CompCRC (Ours)
& None & N/A
& 0.2767 & 0.6767 & 1.7449 & 0.3878 \\
\bottomrule
\end{tabular*}
\vspace{1mm}
\begin{minipage}{0.98\linewidth}
\footnotesize
\emph{Note.} Lower Risk and Cost and higher Acc.\ and
Acc./Cost are preferred. N/A indicates that CRC certification
is Not Applicable. Route-Gated CompCRC is our empirical
variant without a formal CRC guarantee.
\end{minipage}
\endgroup
\end{table}
\FloatBarrier

\paragraph{Comparison with Baselines}
As Table~\ref{tab:main_results} shows, CompCRC achieves
lower observed risk and higher accuracy than Global-CRC
on all three datasets.
On GSM8K and HellaSwag, it also reduces cost and improves
Acc./Cost.
The largest cost saving occurs on HellaSwag, where the
cost ratio decreases from 1.5896 to 1.1069, a reduction
of 30.37\%, while accuracy increases by 2.00 percentage
points.
On MMLU, CompCRC reduces observed risk and improves
accuracy by 1.33 percentage points each, accompanied
by a 6.77\% cost increase and lower Acc./Cost.

Compared with the routing baselines, Route-only and
FrugalGPT-style, CompCRC consistently achieves lower
observed risk and higher accuracy.
Relative to Route-only, these improvements accompany
cost increases of 10.69--28.54\% and lower Acc./Cost.
Relative to FrugalGPT-style, CompCRC improves all four
numerical metrics on GSM8K and HellaSwag, while its
MMLU result incurs higher cost and lower Acc./Cost.

The single-model results provide additional
performance--cost references.
L1 only achieves the lowest cost and highest Acc./Cost
on all three datasets, while CompCRC provides higher
accuracy at additional cost.
Compared with L3 only, CompCRC reduces cost by
approximately 59.0--63.8\%, with lower accuracy.
L2 only achieves higher accuracy at lower cost than
CompCRC on all three datasets.
These results place the certified cascade's performance
alongside the available single-model alternatives and
show that its empirical advantages depend on the comparator.

\begin{figure}[pos=h!]
\centering
\includegraphics[width=\linewidth]{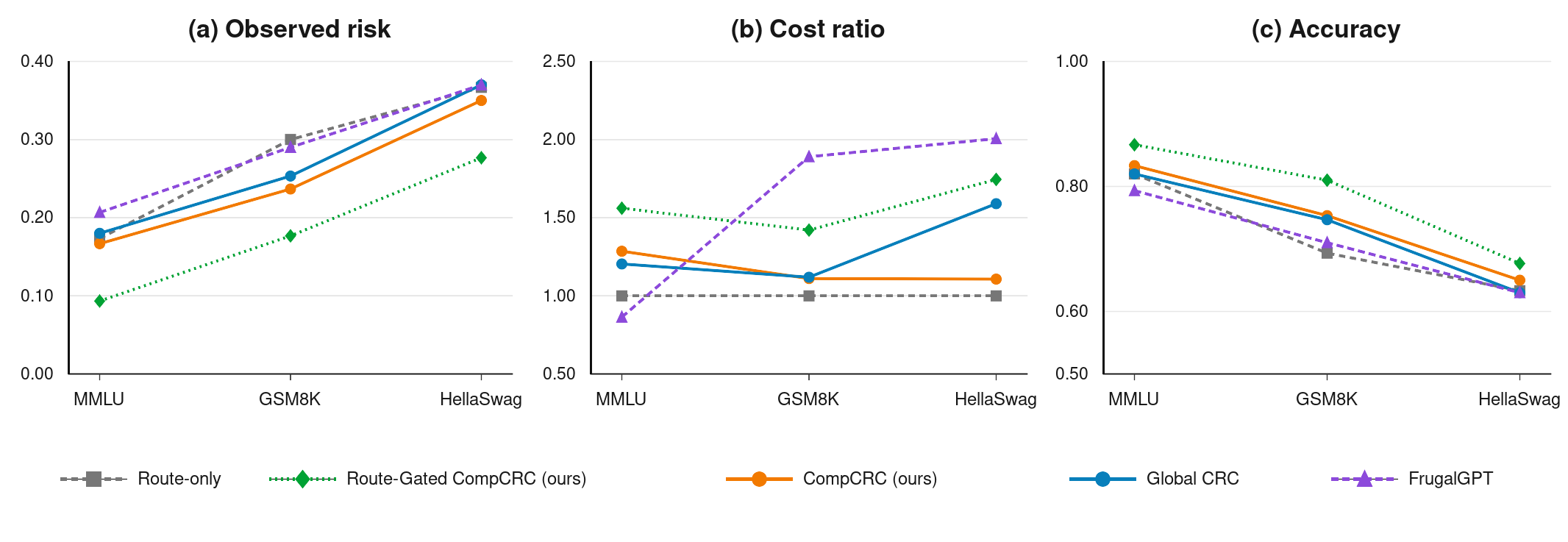}
\caption{Observed risk, cost ratio, and accuracy across the
three datasets at $\alpha=0.45$.}
\label{fig:cross_dataset_results}
\end{figure}

\paragraph{Route-Gated Variant}
Relative to CompCRC, Route-Gated CompCRC further reduces
observed risk and improves accuracy on all three datasets,
with higher cost and lower Acc./Cost.
On MMLU, its observed risk of 0.0933 is the lowest among
all evaluated methods.
These results demonstrate an additional empirical
risk--accuracy--cost trade-off.
Route-Gated CompCRC is evaluated as an empirical variant
and is not covered by the formal guarantee established
for CompCRC.
Figure~\ref{fig:cross_dataset_results} visualizes the
cross-dataset performance comparison.

CompCRC uses the cascade-end fallback for only 3 of 300
GSM8K queries (1.0\%) and none on MMLU or HellaSwag,
so its reported results primarily reflect stage-accepted
outputs.
Additional visualizations of the GSM8K and HellaSwag
results appear in Appendices~\ref{app:gsm8k_fixed}
and~\ref{app:hellaswag_fixed}, respectively.
Appendix~\ref{app:cc_comparison} further compares CompCRC
with three-stage Conformal Cascade (CC-3) on the same
MMLU queries, evaluating final accuracy and weighted
invocation cost under their respective inference
configurations.

\par\vspace{0.5\baselineskip}
Overall, at the reported operating point, CompCRC combines
stage-wise risk certification with lower observed risk
and higher accuracy than Global-CRC and the routing
baselines on all three datasets.
Relative to Global-CRC, it additionally reduces cost
on GSM8K and HellaSwag, while its MMLU improvements
require additional cost.

\FloatBarrier

\subsection{Stage-Wise Certification and Diagnosis}
\label{subsec:stagewise_results}

To answer RQ2, we examine CompCRC's stage-wise certification
results at two MMLU operating points: $\alpha=0.45$, which
passes certification, and $\alpha=0.30$, which is marked NC.
Each operating point uses its own policy, selected and
frozen before certification following
Section~\ref{sec:method}.
Table~\ref{tab:mmlu_certification} reports the local budgets
$\alpha_k$, CRC-corrected statistics $\widetilde z_k$,
budget-utilization ratios $\widetilde z_k/\alpha_k$,
certification results, and frozen acceptance thresholds $t_k$.
A utilization ratio at or below 1 satisfies the local check;
within this range, a higher ratio indicates less relative
budget headroom, while a ratio above 1 identifies a local
budget violation.
The System row summarizes the entire candidate cascade,
reporting the global budget $\alpha$, the summed statistic
$\Sigma=\sum_k\widetilde z_k$, and the overall certification
status.
This status is determined by the conjunction of all local
checks, with $\Sigma$ reported only as a summary.

\paragraph{Certification Success and Budget Headroom}
As Table~\ref{tab:mmlu_certification} shows, at $\alpha=0.45$,
the candidate cascade selected under the cost rule passes
all three local checks, yielding $A=1$.
L1 and L2 each use approximately 98\% of their allocated
budgets, while L3 uses approximately 38\%.
Beyond confirming certification success, the stage-wise
results identify L1 and L2 as operating close to their
local budget limits and L3 as having substantially greater
relative budget headroom.

\paragraph{Certification Failure and Diagnosis}
At $\alpha=0.30$, the corrected statistics for L1 and L2
are approximately 1.95 and 2.62 times their respective
budgets, whereas L3 passes its local check.
The operating point is therefore marked NC ($A=0$),
and the selected cascade is replaced by the predefined
certification-failure fallback policy.
The local checks identify L1 and L2 as the stages whose
budget violations prevent certification.
Such violations can also occur when the summed statistics
fall below the global budget.
On GSM8K at $\alpha=0.25$, $\Sigma=0.1816<0.25$, but L3
exceeds its local budget, so this operating point is also
marked NC.
This case illustrates the additional diagnostic information
provided by the local checks: the summed statistic alone
does not reveal which stage violates its allocated budget.
Further certification results for GSM8K and HellaSwag
are provided in Appendix~\ref{app:cross_dataset_sweeps}.

\par\vspace{0.5\baselineskip}
These findings indicate that CompCRC provides a system-level certification
decision together with stage-wise budget diagnostics,
revealing limited budget headroom at certified operating
points and localizing the budget violations responsible
for certification failure.

\begin{table}[pos=h!]
\caption{Stage-wise certification and budget utilization
on MMLU.}
\label{tab:mmlu_certification}
\centering
\begingroup
\small
\setlength{\tabcolsep}{4pt}
\renewcommand{\arraystretch}{1.10}
\begin{tabular*}{\linewidth}{
@{\extracolsep{\fill}}c l r r r c r@{}}
\toprule
$\alpha$ &
\textbf{Stage} &
$\alpha_k$ &
$\widetilde z_k$ &
$\widetilde z_k/\alpha_k$ &
\textbf{CRC status} &
$t_k$ \\
\midrule
\multirow{4}{*}{0.45}
& L1 & 0.2632 & 0.2571 & 0.98 & $\checkmark$ & 0.8036 \\
& L2 & 0.1316 & 0.1285 & 0.98 & $\checkmark$ & 0.5331 \\
& L3 & 0.0553 & 0.0208 & 0.38 & $\checkmark$ & 0.3127 \\
& System & 0.4500 & $\Sigma=0.4064$ & -- & Pass & -- \\
\midrule
\multirow{4}{*}{0.30}
& L1 & 0.1817 & 0.3542 & 1.95 & $\times$ & 0.7419 \\
& L2 & 0.0814 & 0.2132 & 2.62 & $\times$ & 0.3956 \\
& L3 & 0.0369 & 0.0052 & 0.14 & $\checkmark$ & 0.5049 \\
& System & 0.3000 & $\Sigma=0.5726$ & -- & NC & -- \\
\bottomrule
\end{tabular*}
\vspace{1mm}
\begin{minipage}{0.98\linewidth}
\footnotesize
\emph{Note.} Stage symbols $\checkmark$ and $\times$
indicate passing and failing local checks, respectively.
The System row reports the global budget in the budget
column and the overall status as Pass ($A=1$) or NC ($A=0$).
Budget-utilization ratios are rounded to two decimal places.
\end{minipage}
\endgroup
\end{table}
\FloatBarrier

\subsection{Effect of Risk-Budget Allocation}
\label{subsec:cost_vs_uniform}

To answer RQ3, we compare the independently selected cost rule
with a pre-specified uniform-allocation baseline on three
datasets at the same global risk budget $\alpha=0.45$.
Table~\ref{tab:cost_uniform} reports the results of the frozen
policies on the independent test sets, together with their
highest-utilization stages.
Both allocations pass all stage-wise checks on each dataset.
The Rule column identifies the allocation rule.
Risk and Acc.\ denote observed accepted-error risk and
accuracy, respectively; Cost is the normalized inference
cost, and Acc./Cost measures accuracy per unit of normalized
cost.
Lower Risk and Cost and higher Acc.\ and Acc./Cost are
preferred, with Cost and Acc./Cost serving as the primary
cost-efficiency indicators in this comparison.
The highest-utilization stage maximizes
$\widetilde z_k/\alpha_k$ and identifies the stage with the
tightest relative budget; the stage label itself has no
preferred ordering.

\paragraph{Risk--Accuracy--Cost Trade-offs}
As Table~\ref{tab:cost_uniform} shows, cost-based allocation
reduces the cost ratio and improves Acc./Cost on all three
datasets.
Relative to uniform allocation, Acc./Cost increases by
approximately 16.8\% on MMLU, 10.6\% on GSM8K, and 46.7\%
on HellaSwag.
HellaSwag exhibits the largest improvement, with the cost
ratio decreasing from 1.6573 to 1.1069, a reduction of
approximately 33.21\%.
These cost-efficiency gains are accompanied by accuracy
reductions of 2.67, 3.00, and 1.33 percentage points on
MMLU, GSM8K, and HellaSwag, respectively, together with
higher observed risk.
Thus, the comparison demonstrates a trade-off between
predictive performance and inference cost among configurations
that pass the stage-wise checks.

\paragraph{Stage-Wise Budget Utilization}
Under uniform allocation, the highest-utilization stages
are L2 on MMLU, L1 on GSM8K, and L2 on HellaSwag.
Under cost-based allocation, L1 has the highest utilization
on all three datasets.
On MMLU, however, L1 and L2 have nearly equal utilization
ratios of approximately 98\%, indicating that both stages
operate close to their respective local budgets.
These results show that the allocation rule changes the
relative tightness of stage-wise budget constraints.
The utilization diagnostic complements the aggregate
performance metrics by identifying where each frozen
policy has the least relative budget headroom.

\par\vspace{0.5\baselineskip}
Overall, at the evaluated global budget, cost-based
allocation changes stage-wise budget utilization and
achieves lower inference cost and higher Acc./Cost than
uniform allocation on all three datasets, at the expense
of some accuracy, while satisfying all stage-wise checks.

\begin{table}[pos=h!]
\caption{Comparison of uniform and cost-based risk-budget
allocation at $\alpha=0.45$.}
\label{tab:cost_uniform}
\centering
\begingroup
\small
\setlength{\tabcolsep}{4pt}
\renewcommand{\arraystretch}{1.10}
\begin{tabular*}{\linewidth}{
@{\extracolsep{\fill}}l l c r r r r@{}}
\toprule
\textbf{Dataset} &
\textbf{Rule} &
\shortstack{\textbf{Highest-utilization}\\\textbf{stage}} &
\textbf{Risk}$\downarrow$ &
\textbf{Acc.}$\uparrow$ &
\textbf{Cost}$\downarrow$ &
\textbf{Acc./Cost}$\uparrow$ \\
\midrule
\multirow{2}{*}{MMLU}
& Uniform & L2
& \textbf{0.1400} & \textbf{0.8600}
& 1.5489 & 0.5552 \\
& Cost & L1
& 0.1667 & 0.8333
& \textbf{1.2854} & \textbf{0.6483} \\
\midrule
\multirow{2}{*}{GSM8K}
& Uniform & L1
& \textbf{0.2067} & \textbf{0.7833}
& 1.2775 & 0.6132 \\
& Cost & L1
& 0.2367 & 0.7533
& \textbf{1.1105} & \textbf{0.6783} \\
\midrule
\multirow{2}{*}{HellaSwag}
& Uniform & L2
& \textbf{0.3367} & \textbf{0.6633}
& 1.6573 & 0.4002 \\
& Cost & L1
& 0.3500 & 0.6500
& \textbf{1.1069} & \textbf{0.5872} \\
\bottomrule
\end{tabular*}
\vspace{1mm}
\begin{minipage}{0.98\linewidth}
\footnotesize
\emph{Note.} All reported configurations pass the stage-wise
checks ($A=1$).
Bold values indicate the better result for each performance
metric within each dataset.
Highest-utilization stage is a diagnostic label computed
from the frozen policies and does not affect rule selection
or the certification decision.
\end{minipage}
\endgroup
\end{table}
\FloatBarrier

\subsection{Sensitivity to the Global Risk Budget}
\label{subsec:budget_feasibility}

To answer RQ4, we vary the global risk budget over the
evaluated discrete grid and examine its effects on
certification feasibility, observed risk, and inference cost.
At each budget, CompCRC uses its own candidate policy,
selected and frozen before certification following
Section~\ref{sec:method}.
Figure~\ref{fig:mmlu_budget_sweep} presents the MMLU results;
the corresponding GSM8K and HellaSwag sweeps appear in
Appendix~\ref{app:cross_dataset_sweeps}.
The horizontal axis represents the global risk budget
$\alpha$, with smaller values imposing stricter constraints.
The vertical axes report observed risk and cost ratio,
respectively, both of which are preferred to be lower.
Cost is normalized by the Route-only mean cost.
NC marks indicate operating points that do not pass
certification; their vertical positions do not represent
measured risk or cost.

\begin{figure}[pos=h!]
\centering
\includegraphics[width=\linewidth]{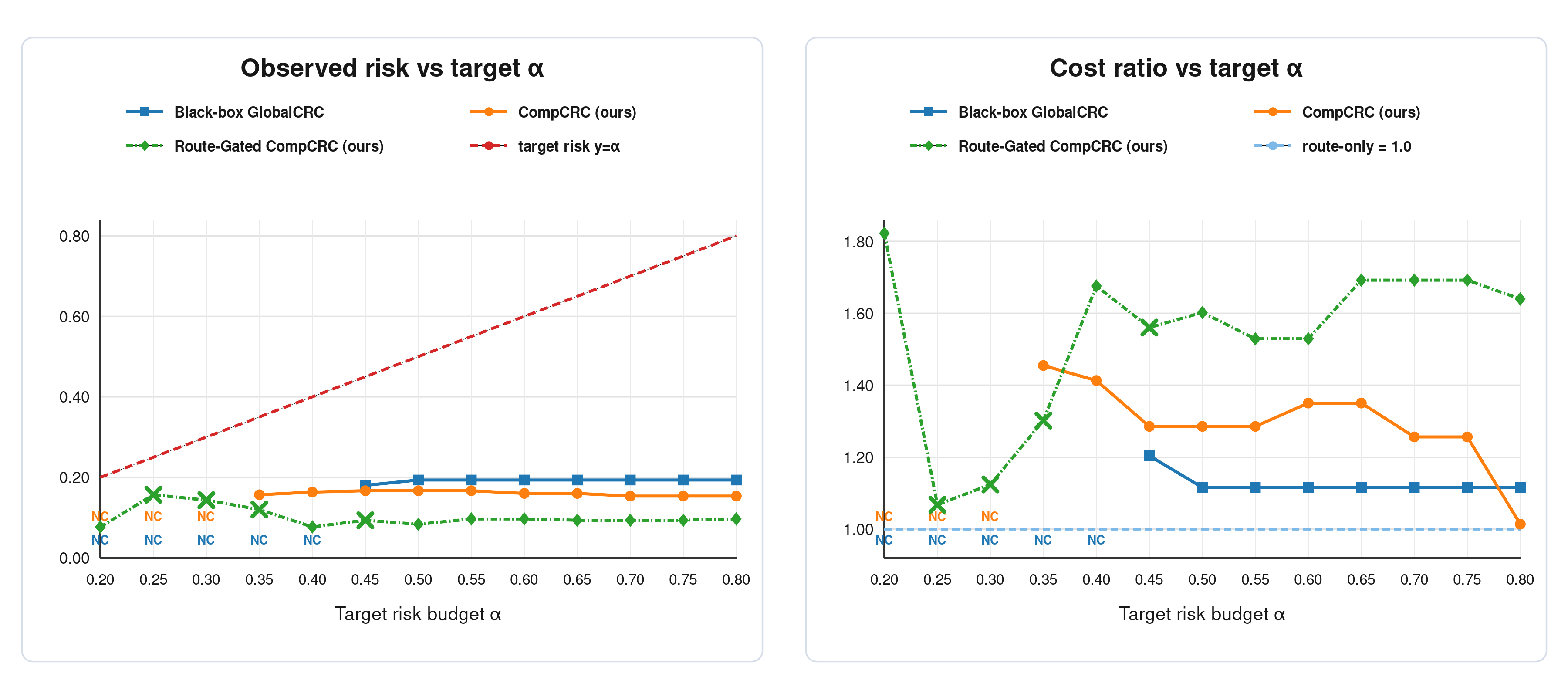}
\caption{Observed risk and cost ratio across the evaluated
MMLU risk budgets. NC marks indicate operating points
that do not pass certification; their vertical positions
do not represent measured risk or cost.}
\label{fig:mmlu_budget_sweep}
\end{figure}
\FloatBarrier

\paragraph{Certification Feasibility across Datasets}
As Figure~\ref{fig:mmlu_budget_sweep} shows, the smallest
evaluated budget at which CompCRC passes certification
on MMLU is $\alpha=0.35$, compared with $\alpha=0.45$
for Global-CRC.
CompCRC therefore passes certification under a stricter
evaluated budget in this comparison.
Its smallest passing budgets on GSM8K and HellaSwag
are $\alpha=0.30$ and $\alpha=0.40$, respectively.
On each dataset, CompCRC passes at every evaluated
budget at or above its corresponding smallest passing value.
These values describe the smallest passing budgets on
the evaluated discrete grid and do not establish
continuous feasibility thresholds.
An NC outcome concerns the candidate selected at that
budget and does not imply that no alternative policy
could pass certification.
Representative stage-wise causes of certification failure
are analyzed in Section~\ref{subsec:stagewise_results}.

\paragraph{Risk--Cost Trends across Budgets}
On MMLU, CompCRC and Global-CRC both pass certification
at $\alpha\in\{0.45,0.50,\ldots,0.80\}$.
Across these common certified operating points,
CompCRC maintains an observed risk of approximately
0.15--0.17, compared with approximately 0.18--0.19
for Global-CRC.
Its lower observed risk therefore extends across
multiple evaluated budgets.
As the budget is relaxed, the CompCRC cost ratio
decreases overall and approaches that of Route-only
at $\alpha=0.80$, although the cost curve is not
strictly monotone.

\par\vspace{0.5\baselineskip}
Taken together, these results show that the global risk
budget affects both certification feasibility and the
empirical risk--cost trade-off.
On MMLU, CompCRC passes certification at a smaller
evaluated budget than Global-CRC and maintains lower
observed risk at their common certified operating points,
while its cost decreases overall as the budget is relaxed.

\Needspace{22\baselineskip}

\section{Conclusion and Future Work}
\label{sec:conclusion}
CompCRC connects system-level accepted-error risk control to explicit stage-wise constraints, making risk budgets and certification failures traceable to individual stages. Separating policy selection from certification allows accuracy and inference cost to guide selection while preserving a finite-sample guarantee on the complete deployment procedure's marginal expected accepted-error risk under conditional exchangeability. Across the evaluated benchmarks, CompCRC achieves lower observed accepted-error risk and higher final accuracy than Global-CRC, while cost-based budget allocation improves accuracy per unit cost over uniform allocation. These findings support stage-wise certification as a practical basis for combining system-level risk guarantees with local risk diagnosis.

Future work will investigate more data-efficient budget allocation
and certification for longer cascades and more diverse tasks, while
maintaining the separation between policy selection and certification.
Extending the framework to distribution shift and temporally dependent
queries will require calibration procedures and theoretical guarantees
suited to these settings. Incorporating measured latency and
token-dependent costs will further align policy selection with
operational requirements. Finally, jointly controlling errors in
stage-accepted and fallback answers is a promising direction toward
risk guarantees that cover all delivered outputs.

\appendix
\renewcommand{\thesection}{\Alph{section}}
\numberwithin{equation}{section}
\numberwithin{theorem}{section}
\numberwithin{figure}{section}
\numberwithin{table}{section}

\section{Appendix}

\label{app:supplementary_results}

\subsection{GSM8K Results at the Common Risk Budget}
\label{app:gsm8k_fixed}

\begin{figure}[pos=h!]
\centering
\includegraphics[width=\linewidth]{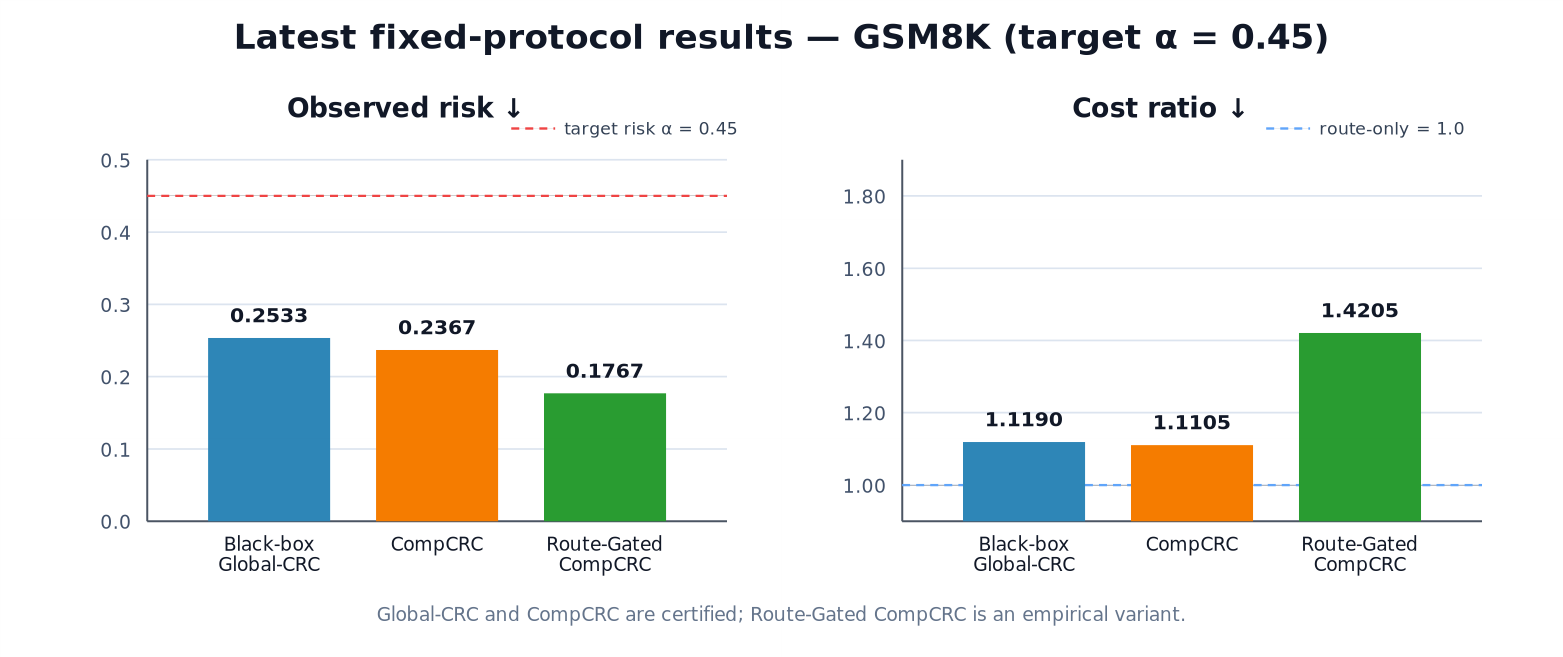}
\caption{Observed risk and cost ratio on GSM8K at
$\alpha=0.45$.}
\label{fig:gsm8k_fixed}
\end{figure}

Figure~\ref{fig:gsm8k_fixed} supplements the GSM8K results
in Section~\ref{subsec:main_results} at $\alpha=0.45$.
CompCRC achieves lower observed risk, higher accuracy,
and slightly lower cost than Global-CRC.
Both methods pass their respective certification procedures.
Our Route-Gated CompCRC variant further reduces observed
risk and improves accuracy, with higher invocation cost.

\FloatBarrier

\subsection{HellaSwag Results at the Common Risk Budget}
\label{app:hellaswag_fixed}

\begin{figure}[pos=h!]
\centering
\includegraphics[width=\linewidth]{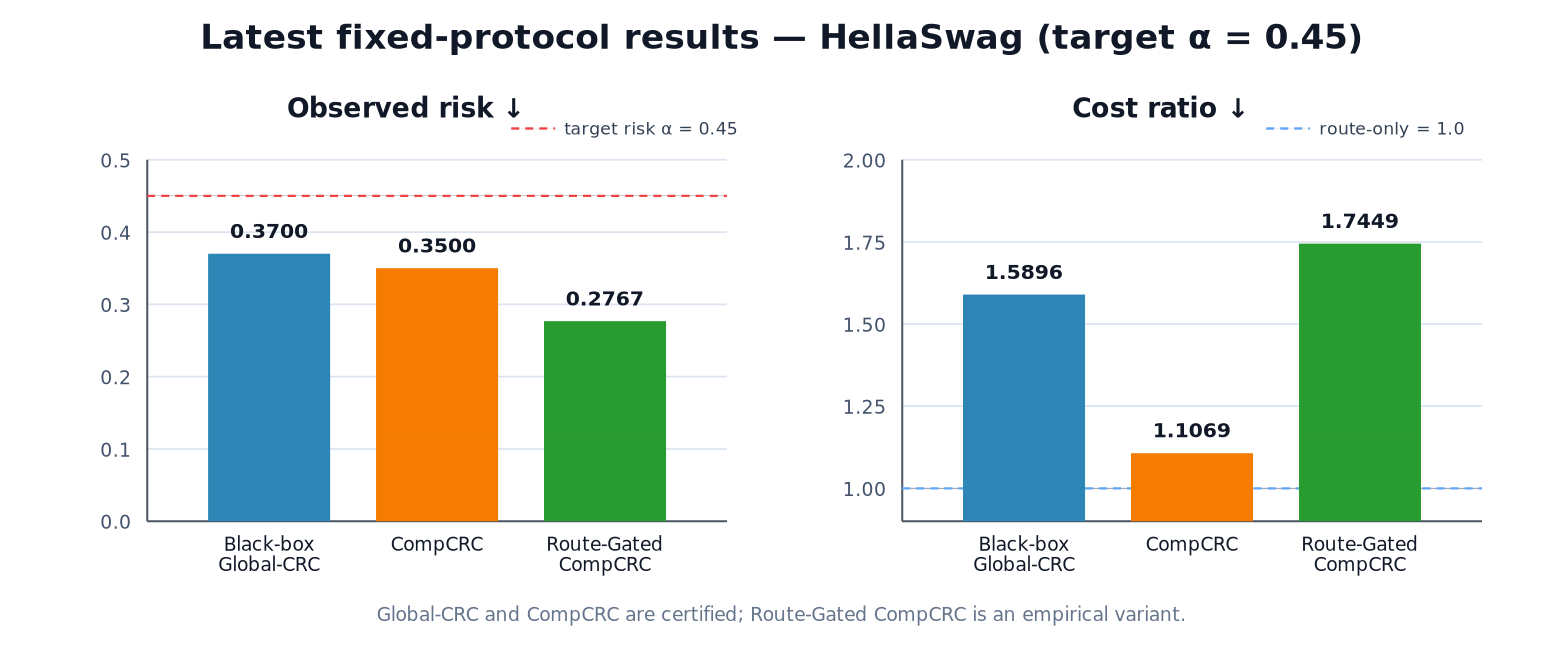}
\caption{Observed risk and cost ratio on HellaSwag at
$\alpha=0.45$.}
\label{fig:hellaswag_fixed}
\end{figure}

Figure~\ref{fig:hellaswag_fixed} supplements the HellaSwag
results at the same risk budget.
CompCRC reduces cost by 30.37\% relative to Global-CRC
while also achieving lower observed risk and higher accuracy.
Its cost is also lower than that of our Route-Gated CompCRC
variant, which further improves risk and accuracy.
Global-CRC and CompCRC pass certification on this dataset;
Route-Gated CompCRC is evaluated as an empirical variant.

\FloatBarrier

\subsection{Comparison with Conformal Cascade on MMLU}
\label{app:cc_comparison}

We compare CompCRC with three-stage Conformal Cascade
(CC-3) on the same 300 MMLU test queries using the same
three-model configuration.
CompCRC uses the frozen cost-based policy selected at
$\alpha=0.45$. CC-3 uses the prediction-set-size acceptance
parameter $\kappa=1$ and stage-wise miscoverage level
$\beta_{\mathrm{CC}}=0.45$.
These parameters control different statistical targets.
The comparison therefore evaluates final answer accuracy
and end-to-end invocation cost under the specified
inference configurations.
CC-3 uses 16 frequency samples per stage at a temperature
of 0.7, whereas CompCRC uses one generation per visited
stage. Weighted invocation cost accounts for all generated
samples using the corresponding stage weights.
Final accuracy includes fallback outputs, and fallback
rate records the proportion of queries entering each
method's designated fallback branch.

\begin{table}[pos=h!]
\caption{End-to-end performance of CC-3 and CompCRC on MMLU.}
\label{tab:cc_compcrc}
\centering
\begingroup
\begin{minipage}{0.82\linewidth}
\small
\setlength{\tabcolsep}{6pt}
\renewcommand{\arraystretch}{1.08}
\begin{tabularx}{\linewidth}{@{}Xcc@{}}
\toprule
\textbf{Metric} &
\textbf{CC-3} &
\textbf{CompCRC (Ours)} \\
\midrule
Final accuracy$\uparrow$
& 85.00\% & 83.33\% \\
Average weighted invocation cost$\downarrow$
& 34.5905 & 1.9514 \\
Fallback rate
& 13.67\% & 0.00\% \\
\bottomrule
\end{tabularx}
\vspace{1mm}
\footnotesize
\emph{Note.} Costs are mean weighted invocation costs per
test query, with one L1 generation assigned unit cost,
before the Route-only normalization used in
Table~\ref{tab:main_results}.
Fallback rates refer to each method's own fallback branch.
\end{minipage}
\endgroup
\end{table}
\FloatBarrier

As shown in Table~\ref{tab:cc_compcrc}, CompCRC achieves
83.33\% final accuracy compared with 85.00\% for CC-3,
a difference of 1.67 percentage points, while reducing
average weighted invocation cost by 94.36\%.
CompCRC accepts all test queries through normal cascade
stages, whereas CC-3 uses fallback for 13.67\% of queries.
CompCRC's stage-wise certification results are reported
in Section~\ref{subsec:stagewise_results}.
The end-to-end cost saving reflects both the cascade
paths and the different sampling requirements of the
two inference procedures.

\subsection{Risk--Cost Sweeps and Certification Results}
\label{app:cross_dataset_sweeps}

\begin{figure}[pos=h!]
\centering
\includegraphics[width=\linewidth]{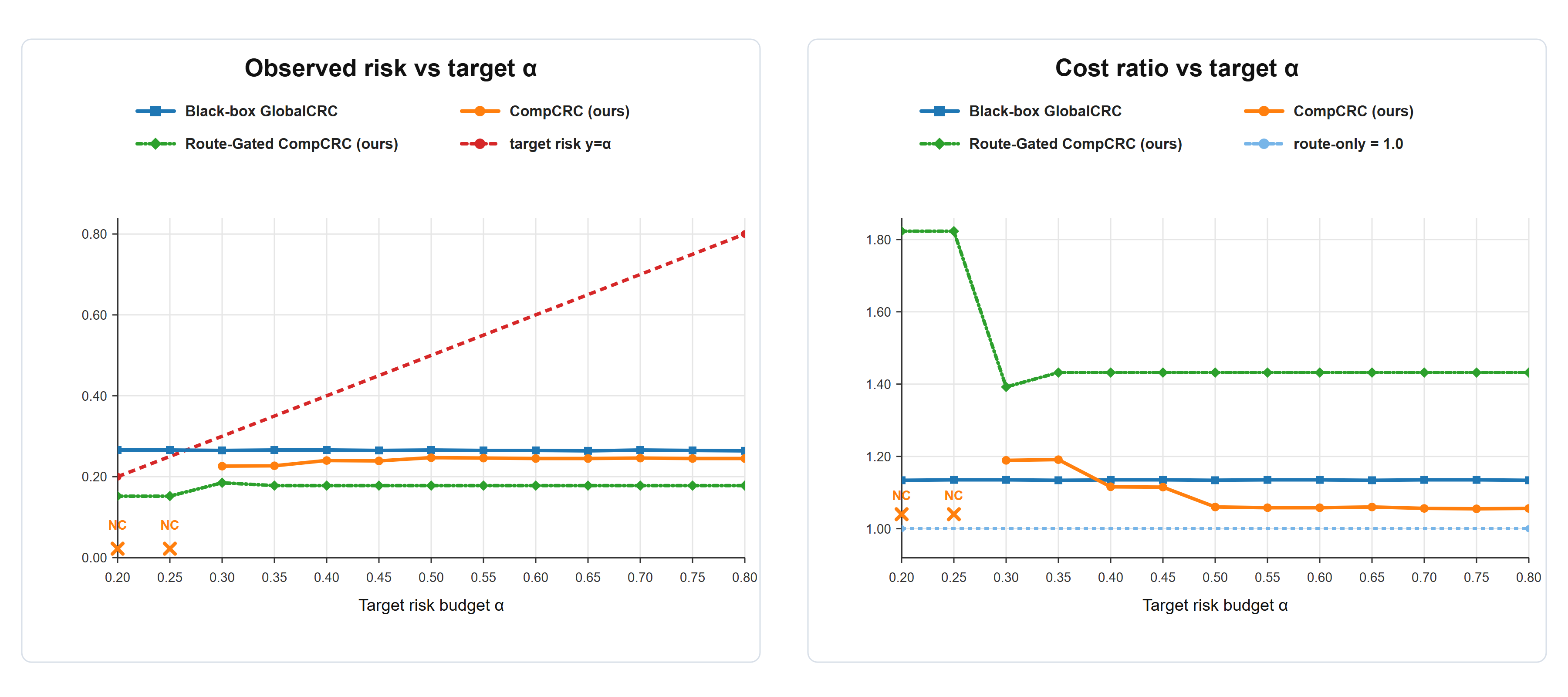}
\caption{Observed risk and cost ratio across the evaluated
GSM8K risk budgets. NC marks indicate operating points
that do not pass certification.}
\label{fig:gsm8k_sweep}
\end{figure}

\begin{figure}[pos=h!]
\centering
\includegraphics[width=\linewidth]{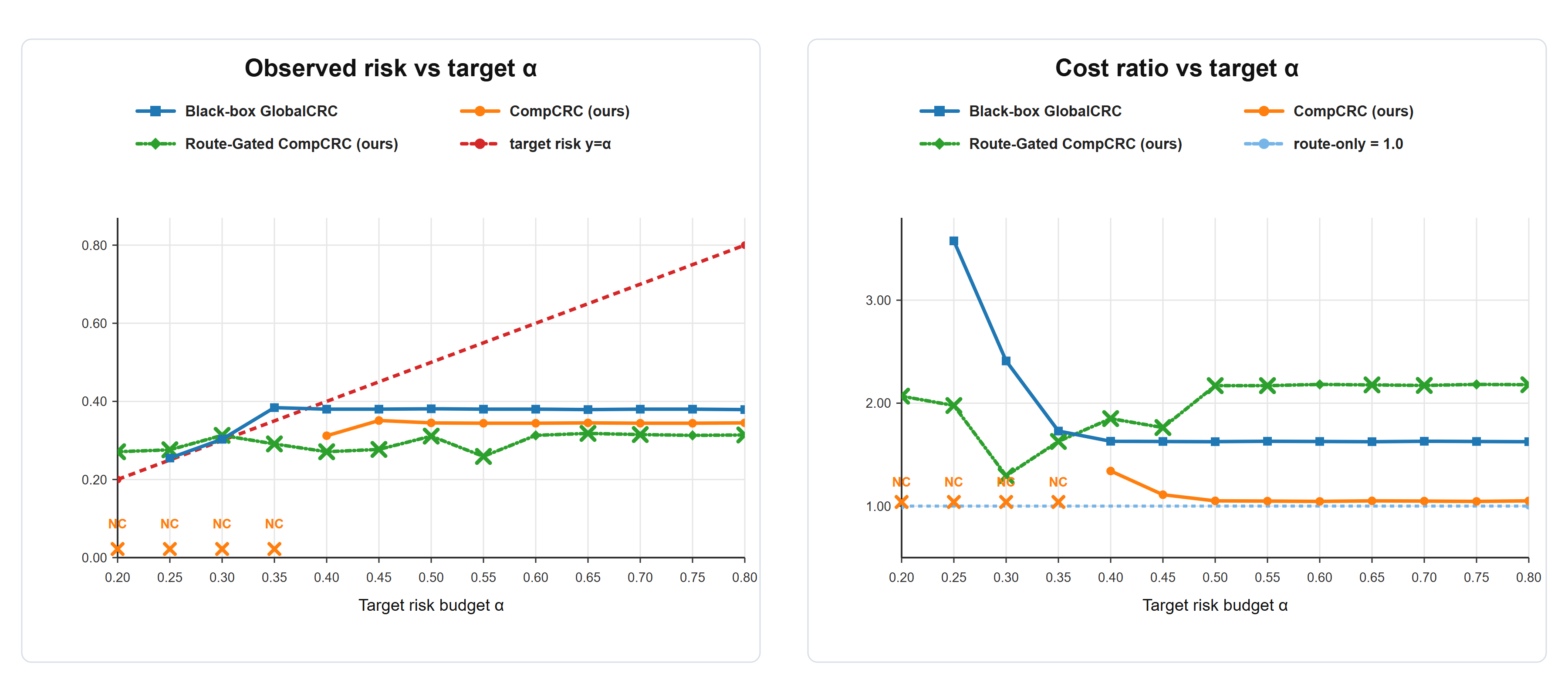}
\caption{Observed risk and cost ratio across the evaluated
HellaSwag risk budgets. NC marks indicate operating points
that do not pass certification.}
\label{fig:hellaswag_sweep}
\end{figure}

Figures~\ref{fig:gsm8k_sweep} and~\ref{fig:hellaswag_sweep}
supplement the MMLU sweep in
Figure~\ref{fig:mmlu_budget_sweep}, using the same discrete
budget grid and cost normalization.
NC marks indicate operating points with $A=0$; their
vertical positions do not represent measured risk or cost.
Tables~\ref{tab:gsm8k_crc_appendix}
and~\ref{tab:hellaswag_crc_appendix} provide stage-wise
results for a non-certified and a certified operating
point on each dataset, following the notation in
Section~\ref{subsec:stagewise_results}.
The System rows report the global budget, the summed
statistic $\Sigma$, and the overall certification status:
Pass ($A=1$) or NC ($A=0$).
Each candidate policy and its local budgets are selected
and frozen before certification.
Test risks reported for NC points describe the candidates
for diagnostic purposes; deployment uses the predefined
certification-failure fallback policy.

\paragraph{GSM8K}
On the evaluated grid, CompCRC is marked NC at
$\alpha=0.20$ and $0.25$ and passes certification at
every evaluated budget with $\alpha\geq0.30$
(Figure~\ref{fig:gsm8k_sweep}).
At $\alpha=0.25$, $\Sigma=0.1816$ is below the global
budget, but L3 exceeds its local budget
($0.0312>0.0307$), with a budget-utilization ratio of
approximately 1.02.
The local checks therefore identify L3 as the stage
preventing certification.
At $\alpha=0.45$, all three stages pass, with L1
showing the highest budget utilization, approximately
66\% (Table~\ref{tab:gsm8k_crc_appendix}).

\FloatBarrier

\begin{table}[pos=h!]
\caption{Stage-wise certification and budget utilization
on GSM8K.}
\label{tab:gsm8k_crc_appendix}
\centering
\begingroup
\small
\setlength{\tabcolsep}{4pt}
\renewcommand{\arraystretch}{1.10}
\begin{tabular*}{\linewidth}{
@{\extracolsep{\fill}}c l r r r c r@{}}
\toprule
$\alpha$ &
\textbf{Stage} &
$\alpha_k$ &
$\widetilde z_k$ &
$\widetilde z_k/\alpha_k$ &
\textbf{CRC status} &
$t_k$ \\
\midrule
\multirow{4}{*}{0.25}
& L1 & 0.1462 & 0.0972 & 0.66 & $\checkmark$ & 0.7143 \\
& L2 & 0.0731 & 0.0533 & 0.73 & $\checkmark$ & 0.6429 \\
& L3 & 0.0307 & 0.0312 & 1.02 & $\times$ & 0.4074 \\
& System & 0.2500 & $\Sigma=0.1816$ & -- & NC & -- \\
\midrule
\multirow{4}{*}{0.45}
& L1 & 0.2632 & 0.1724 & 0.66 & $\checkmark$ & 0.3571 \\
& L2 & 0.1316 & 0.0219 & 0.17 & $\checkmark$ & 0.6429 \\
& L3 & 0.0553 & 0.0234 & 0.42 & $\checkmark$ & 0.4074 \\
& System & 0.4500 & $\Sigma=0.2177$ & -- & Pass & -- \\
\bottomrule
\end{tabular*}
\vspace{1mm}
\begin{minipage}{0.98\linewidth}
\footnotesize
\emph{Note.} Both operating points use the cost rule.
Their candidate test risks are 0.2067 at $\alpha=0.25$
and 0.2367 at $\alpha=0.45$.
Stage symbols $\checkmark$ and $\times$ indicate passing
and failing local checks, respectively.
The System status is the conjunction of all local checks;
$\Sigma$ is reported only as a summary.
Budget-utilization ratios are shown to two decimal places.
\end{minipage}
\endgroup
\end{table}
\FloatBarrier

\paragraph{HellaSwag}
CompCRC is marked NC at every evaluated budget from
$\alpha=0.20$ through $0.35$ and passes certification
at all evaluated budgets with $\alpha\geq0.40$
(Figure~\ref{fig:hellaswag_sweep}).
At $\alpha=0.35$, $\Sigma=0.3185$ is below the global
budget, but L2 exceeds its local budget
($0.1097>0.0880$), with a budget-utilization ratio of
approximately 1.25.
L1 and L3 pass their local checks, identifying L2
as the stage responsible for the NC outcome.
At $\alpha=0.45$, all three stages pass
(Table~\ref{tab:hellaswag_crc_appendix}).

\FloatBarrier

\begin{table}[pos=h!]
\caption{Stage-wise certification and budget utilization
on HellaSwag.}
\label{tab:hellaswag_crc_appendix}
\centering
\begingroup
\small
\setlength{\tabcolsep}{4pt}
\renewcommand{\arraystretch}{1.10}
\begin{tabular*}{\linewidth}{
@{\extracolsep{\fill}}c l r r r c r@{}}
\toprule
$\alpha$ &
\textbf{Stage} &
$\alpha_k$ &
$\widetilde z_k$ &
$\widetilde z_k/\alpha_k$ &
\textbf{CRC status} &
$t_k$ \\
\midrule
\multirow{4}{*}{0.35}
& L1 & 0.2069 & 0.1724 & 0.83 & $\checkmark$ & 0.4564 \\
& L2 & 0.0880 & 0.1097 & 1.25 & $\times$ & 0.6682 \\
& L3 & 0.0551 & 0.0364 & 0.66 & $\checkmark$ & 0.9152 \\
& System & 0.3500 & $\Sigma=0.3185$ & -- & NC & -- \\
\midrule
\multirow{4}{*}{0.45}
& L1 & 0.2632 & 0.2257 & 0.86 & $\checkmark$ & 0.4206 \\
& L2 & 0.1316 & 0.1066 & 0.81 & $\checkmark$ & 0.5472 \\
& L3 & 0.0553 & 0.0234 & 0.42 & $\checkmark$ & 0.4243 \\
& System & 0.4500 & $\Sigma=0.3557$ & -- & Pass & -- \\
\bottomrule
\end{tabular*}
\vspace{1mm}
\begin{minipage}{0.98\linewidth}
\footnotesize
\emph{Note.} The $\alpha=0.35$ operating point uses the
cost--error rule, and the $\alpha=0.45$ operating point
uses the cost rule.
Their candidate test risks are 0.3000 and 0.3500,
respectively.
Stage symbols $\checkmark$ and $\times$ indicate passing
and failing local checks, respectively.
The System status is the conjunction of all local checks;
$\Sigma$ is reported only as a summary.
Budget-utilization ratios are shown to two decimal places.
\end{minipage}
\endgroup
\end{table}
\FloatBarrier

These results show how CompCRC localizes certification
failures even when $\Sigma<\alpha$.
The stage-wise statistics identify both the stage
preventing certification and the extent of its local
budget violation, providing a direct explanation
of the overall NC outcome.

\bibliography{references}
\end{document}